\documentclass{article}
\PassOptionsToPackage{numbers, compress}{natbib}

\usepackage[final]{neurips_2021}

\usepackage[utf8]{inputenc} 
\usepackage[T1]{fontenc}    
\usepackage{hyperref}       
\usepackage{url}            
\usepackage{booktabs}       
\usepackage{amsfonts}       
\usepackage{nicefrac}       
\usepackage{microtype}      
\usepackage{xcolor}         
\usepackage{subcaption}
\usepackage{float}
\usepackage{graphicx}
\usepackage{xspace}
\usepackage{wrapfig}
\usepackage{amsmath}

\include{figures}

\title{Instance-Conditioned GAN}

\author{
  Arantxa Casanova \\
  Facebook AI Research \\
  École Polytechnique de Montréal \\
Mila, Quebec AI Institute \\
  \And
  Marlène Careil \\
  Facebook AI Research \\
  Télécom Paris \\
      \And
  Jakob Verbeek \\
  Facebook AI Research \\
         \And
  Michał Drożdżal$^*$
 \\
  Facebook AI Research \\

            \And
  Adriana Romero-Soriano\thanks{Equal contribution.}
 \\
  Facebook AI Research \\
  McGill University \\

}

\def \ours {{IC-GAN}\xspace}
\def \ImNet {ImageNet\xspace}
\def \COCO {COCO-Stuff\xspace}
\def\myres#1{$#1\!\times\!#1$}

\def \eg {\textit{e.g.}\xspace}

\def \vs {\textit{vs.}\xspace}
\def \wrt {{w.r.t.}\xspace}

\def\be{\begin{equation}}
\def\ee{\end{equation}}
\def\bea{\begin{eqnarray}}
\def\eea{\end{eqnarray}}
\def\fig#1{Figure~\ref{fig:#1}}

\def\sect#1{Section~\ref{sec:#1}}

\begin{document}
\maketitle
\begin{abstract}

Generative Adversarial Networks (GANs) can generate near photo realistic images  in narrow  domains such as human faces. Yet, modeling complex distributions of datasets such as \ImNet and \COCO remains challenging in unconditional settings.
In this paper, we take inspiration from kernel density estimation techniques and introduce a non-parametric approach to modeling distributions of complex datasets. 
We partition the data manifold into a mixture of overlapping neighborhoods described by a datapoint and its nearest neighbors, and introduce a model, called instance-conditioned GAN (\ours), which learns the distribution around each datapoint. Experimental results on \ImNet and COCO-Stuff show that \ours significantly improves over unconditional models and unsupervised data partitioning baselines. Moreover, we show that  \ours can effortlessly transfer to datasets not seen during training by simply changing the conditioning instances, and still generate realistic images. Finally, we extend \ours to the class-conditional case and show semantically controllable generation and competitive quantitative results on \ImNet; while improving over BigGAN  on \ImNet-LT. 
Code and trained models  to reproduce the reported results are available at \url{https://github.com/facebookresearch/ic_gan}.
\looseness-1
\end{abstract}

\section{Introduction}
Generative Adversarial Networks (GANs)~\cite{goodfellow2014generative} have  shown impressive results in unconditional image generation~\cite{karras2019stylebased, Karras2019stylegan2}. 
Despite their success, GANs present optimization difficulties and can suffer from mode collapse, resulting in the generator not being able to obtain a good distribution coverage, and often producing poor quality and/or low diversity generated samples. 
Although many approaches attempt to mitigate this problem -- \eg~\cite{gulrajani2017improved,lin2018pacgan,lucas18icml,metz2016unrolled} --, complex data distributions such as the one in ImageNet~\cite{ILSVRC15} remain a challenge for unconditional GANs~\cite{liu2020diverse,pmlr-v97-lucic19a}. 
Class-conditional GANs~\cite{brock2018large,mirza2014conditional, miyato2018spectral, zhang2019self} ease the task of learning the data distribution by conditioning on class labels, effectively partitioning the data. Although they provide higher quality samples than their unconditional counterparts, they require labelled data, which may be unavailable or costly to obtain.

Several recent approaches explore the use of unsupervised data partitioning to improve GANs~\cite{armandpour2021partition,eghbal2019mixture,Ghosh_2018_CVPR,hoang2018mgan,liu2020diverse,noroozi2020self}.
While these methods are promising and yield
visually appealing samples, their quality is still far from those obtained with class-conditional GANs. 
These methods make use of relatively coarse and non-overlapping data partitions, which oftentimes contain data points
from different types of objects or scenes.
This diversity of data points may result in a manifold with low density regions, which degrades the quality of the generated samples~\cite{devries2020instance}. 
Using finer partitions, however, tends to deteriorate  results~\cite{liu2020diverse,pmlr-v97-lucic19a,noroozi2020self} because the clusters may contain too few data points for the generator and discriminator to properly model their data distribution.

\begin{figure}[t!]
\centering
\begin{subfigure}[t]{0.48\textwidth}
 \centering
\includegraphics[width=\textwidth]{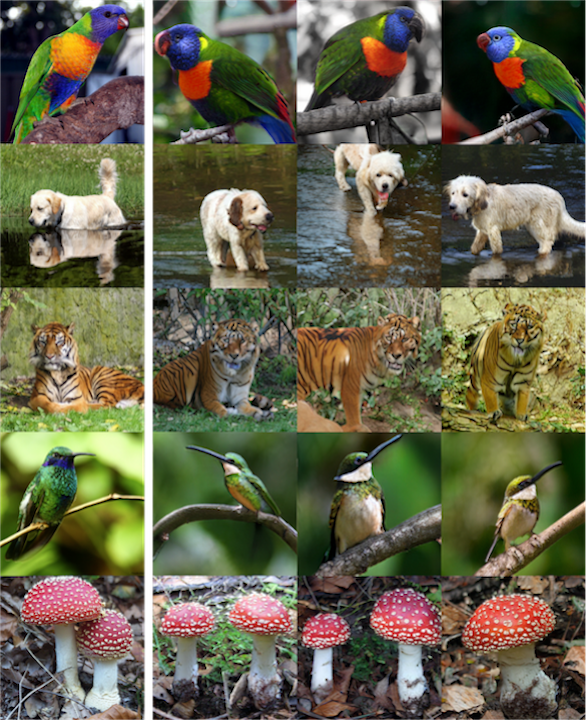}
\caption{\ours samples}
\label{subfig:teaser_unsup}
\end{subfigure}
\begin{subfigure}[t]{0.48\textwidth}
 \centering
\includegraphics[width=\textwidth]{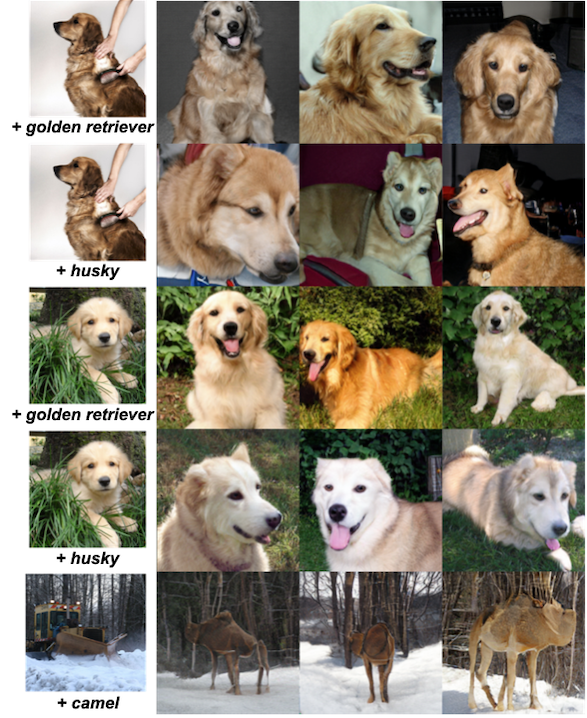}
\caption{Class-conditional \ours samples}
\label{subfig:teaser_class}
\end{subfigure}\\
\begin{subfigure}[t]{0.48\textwidth}
 \centering
\includegraphics[width=\textwidth]{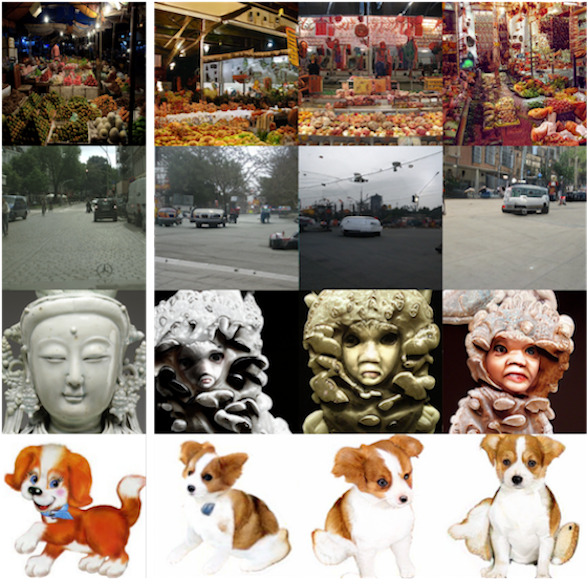}
\caption{\ours transfer samples}
\label{subfig:teaser_unsup_transfer}
\end{subfigure}
\begin{subfigure}[t]{0.48\textwidth}
 \centering
\includegraphics[width=\textwidth]{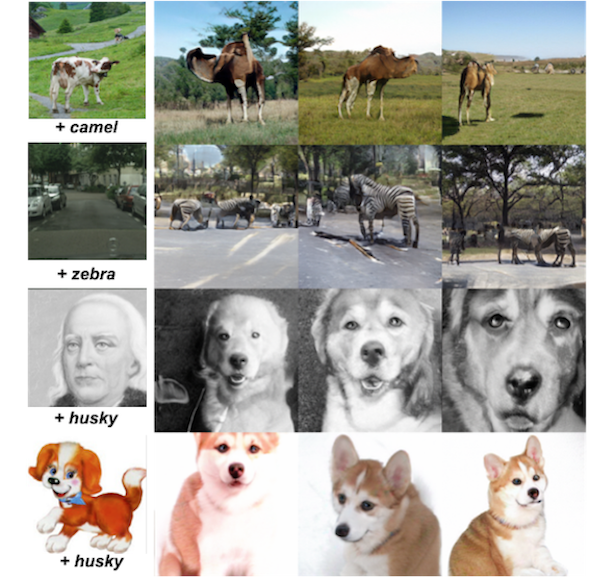}
\caption{Class-conditional \ours transfer samples}
\label{subfig:teaser_class_transfer}
\end{subfigure}

\caption{Samples from unlabeled (a) and class-conditional (b) \ours trained on  the $256\!\times\!256$ \ImNet dataset. For each subfigure, the first column represents instances used to condition the model and the next three columns depict model samples. 
For class-conditional generation in (b) we include samples conditioned  on the same image but different labels.
We highlight the generalization capacities of \ours by applying the \ImNet-trained model to instances from other datasets in unlabeled (c) and class-conditional (d) scenarios. 
Panels (c) and (d) display  samples conditioned on instances from the COCO-Stuff, Cityscapes, MetFaces, and PACS  datasets (from top to bottom). 
}

\label{fig:transfer_qualitative}
\vspace{-0.5cm}
\end{figure}

In this work, we introduce a new approach, called instance-conditioned GAN (\ours), which extends the GAN framework to model a mixture of local data densities.
More precisely, \ours learns to model the distribution of the neighborhood of a data point, also referred to as \textit{instance}, by providing a representation of the instance as an additional input to both the generator and discriminator, and by using the neighbors of the instance as \emph{real} samples for the discriminator.
By choosing a sufficiently large neighborhood around the conditioning instance, we avoid the pitfall of excessively partitioning the data into small clusters. Given the overlapping nature of these clusters, increasing the number of partitions does not come at the expense of having less samples in each of them.
Moreover, unlike when conditioning on discrete cluster indices, conditioning on instance representations naturally leads the generator to produce similar samples for similar instances. 
Interestingly, once trained, our \ours can be used to effortlessly transfer to other datasets not seen during training by simply swapping-out the conditioning instances at inference time. 

\ours bears  similarities with kernel density estimation (KDE), a non-parametric density estimator in the form of a mixture of parametrized kernels modeling the density around each training data point -- see \eg~\cite{bishop06patrec}.
Similar to KDE, \ours can be seen as a mixture density estimator, where each component is obtained by conditioning on a training instance. %
Unlike KDE, however, we do not model the data likelihood explicitly, but take an adversarial approach in which we model the local density implicitly with a neural network that takes as input the conditioning instance as well as a noise vector. Therefore, the \emph{kernel} in \ours is no longer independent on the data point on which we condition, and 
instead of a kernel bandwidth parameter, we control the smoothness by choosing the neighborhood size of an instance from which we sample the \emph{real} samples to be fed to the discriminator.\looseness-1%

We validate our approach on two image generation tasks: (1) \emph{unlabeled} image generation where there is no class information available, and (2) \emph{class-conditional} image generation. For the unlabeled scenario, we report results on the \ImNet and COCO-Stuff datasets. We show that \ours outperforms  previous approaches in unlabeled image generation on both datasets. Additionally, we perform a series of transfer experiments and demonstrate that an \ours trained on \ImNet achieves better generation quality and diversity when testing on COCO-Stuff than the same model trained on COCO-Stuff. 
In the class-conditional setting, we show that \ours can generate images with controllable semantics -- by adapting both class and instance--, while achieving competitive sample quality and diversity on the \ImNet dataset. Finally, we test \ours in \ImNet-LT, a long-tail class distribution ablated version of \ImNet, highlighting the benefits of non-parametric density estimation in datasets with unbalanced classes. \fig{transfer_qualitative} shows \ours unlabeled \ImNet generations (a), \ours class-conditional \ImNet generations (b), and \ours transfer generations both in the unlabeled (c) and controllable class-conditional (d) setting.

\section{Instance-conditioned GAN}

The key idea of \ours is to model the distribution of a complex dataset by leveraging fine-grained overlapping clusters in the data manifold, where each cluster is described by a datapoint $\mathbf{x}_i$ -- referred to as \emph{instance} -- and its nearest neighbors set $\mathcal{A}_i$ in a feature space. Our objective is to model the underlying data distribution $p(\mathbf{x})$ as a mixture of \textit{conditional distributions} $p(\mathbf{x}|\mathbf{h}_i)$ around each of $M$ instance feature vectors $\mathbf{h}_i$ in the dataset, such that $p(\mathbf{x}) \approx \frac{1}{M}\sum_i p(\mathbf{x}|\mathbf{h}_i)$.

\begin{figure}
\centering
\begin{subfigure}[b]{0.35\textwidth}
 \centering
\includegraphics[width=\textwidth]{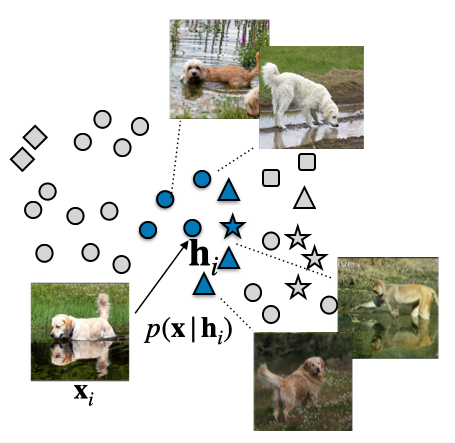}
\caption{
Neighborhood $\mathcal{A}_i$ of instance $\mathbf{h}_i$
}
\label{subfig:example_neighborhood}
\end{subfigure}
\quad\quad\quad
\begin{subfigure}[b]{0.55\textwidth}
 \centering
\includegraphics[width=\textwidth]{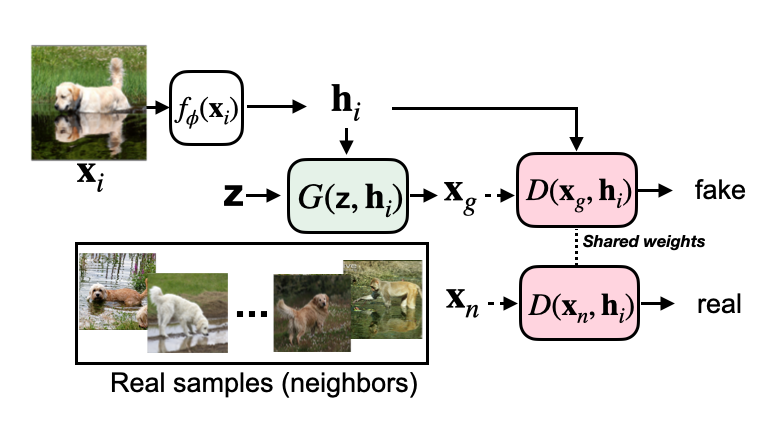}
\caption{Schematic illustration of the \ours workflow}
\label{subfig:icgan_workflow}
\end{subfigure}
\caption{Overview of \ours. (a) The goal of the  generator is to generate realistic images similar to the neighbors of $\mathbf{h}_i$, defined in the embedding space using cosine similarity. Five out of seven neighbors are shown in the figure. Note that images in the same neighborhood may belong to different classes (depicted as different shapes). 
(b) Conditioned on  instance features $\mathbf{h}_i$  and noise $\mathbf{z}$, the generator produces a synthetic sample $\mathbf{x}_g$. 
 Generated samples and real samples (neighbors of $\mathbf{h}_i$) are fed to the discriminator, which is  conditioned on the same $\mathbf{h}_i$.
 }

\label{fig:main_figure}
\end{figure}

More precisely, given an unlabeled dataset $\mathcal{D} = \{\mathbf{x}_i\}_{i=1}^M$ with $M$ data samples $\mathbf{x}_i$ and an embedding function $f$ parametrized by $\phi$, we start by extracting instance features $\mathbf{h}_i=f_{\phi}(\mathbf{x}_i) \; \forall \mathbf{x}_i \in \mathcal{D}$, where $f_{\phi}(\cdot)$ is learned in an unsupervised or self-supervised manner. We then define the set $\mathcal{A}_i$ of $k$ nearest neighbors for each data sample using the cosine similarity -- as is common in nearest neighbor classifiers, \eg~\cite{touvron20arxiv,vinyals16nips} -- over the features $\mathbf{h}_i$. Figure~\ref{subfig:example_neighborhood} depicts a sample $\mathbf{x}_i$ and its  nearest neighbors. \looseness-1

We are interested in implicitly modelling the conditional distributions $p(\mathbf{x}|\mathbf{h}_i)$ with a generator $G_{\theta_G}(\mathbf{z}, \mathbf{h}_i)$, implemented by a deep neural network with parameters $\theta_G$. The generator transforms samples from a unit Gaussian prior $\mathbf{z}\sim \mathcal{N}(0,I)$ into samples $\mathbf{x}$ from the conditional distribution $p(\mathbf{x}|\mathbf{h}_i)$, where $\mathbf{h}_i$ is the feature vector of an instance $\mathbf{x}_i$ sampled from the training data. In \ours, we adopt an adversarial approach to train the generator $G_{\theta_G}$. Therefore, our generator is jointly trained with a discriminator $D_{\theta_D}(\mathbf{x}, \mathbf{h}_i)$ that discerns between real neighbors and generated neighbors of $\mathbf{h}_i$, as shown in Figure~\ref{subfig:icgan_workflow}.
Note that for each $\mathbf{h}_i$, real neighbors are sampled uniformly from $\mathcal{A}_i$.

Both $G$ and $D$ engage in a two player min-max game where they try to find the Nash equilibrium for the following equation:
\begin{equation}
\begin{split}
\min_{G}\max_{D} \; &  \mathbb{E}_{\mathbf{x}_i\sim p(\mathbf{x}), \mathbf{x}_n\sim \mathcal{U}(\mathcal{A}_i)} [\log D(\mathbf{x}_n, f_{\phi}(\mathbf{x}_i))] \; + \\ & \hspace{0.5em} \mathbb{E}_{ \mathbf{x}_i\sim p(\mathbf{x}),\mathbf{z}\sim p(\mathbf{z}) } [\log(1-D(G(\mathbf{z}, f_{\phi}(\mathbf{x}_i)), f_{\phi}(\mathbf{x}_i)))].
\end{split}
\end{equation}

Note that when training \ours we use all available training datapoints to condition the model. At inference time, as in non-parametric density estimation methods such as KDE, the generator of \ours also requires instance features, which may come from the training distribution or a different one. \looseness-1 

\textbf{Extension to class-conditional generation.}
We extend \ours for class-conditional generation by additionally conditioning the generator and discriminator on a class label $\mathbf{y}$. 
More precisely, given a labeled dataset $\mathcal{D}_l = \{(\mathbf{x}_i, \mathbf{y}_i)\}_{i=1}^M$ with $M$ data sample pairs $(\mathbf{x}_i, \mathbf{y}_i)$ and an embedding function $f_\phi$, we  extract instance features $\mathbf{h}_i=f_{\phi}(\mathbf{x}_i) \; \forall \mathbf{x}_i \in \mathcal{D}_l$, where $f_{\phi}(\cdot)$ is learned in an unsupervised, self-supervised, or supervised manner. We then define the set $\mathcal{A}_i$ of $k$ nearest neighbors for each data sample using the cosine similarity over the features $\mathbf{h}_i$, where 
neighbors may be from different classes. This results in neighborhoods, where the number of neighbors belonging to the same class as the instance $\mathbf{h}_i$ is often smaller than $k$. During training, real neighbors $\mathbf{x}_j$ and their respective labels $\mathbf{y}_j$ are sampled uniformly from $\mathcal{A}_i$ for each $\mathbf{h}_i$.
In the class-conditional case, we model $p(\mathbf{x}|\mathbf{h}_i, \mathbf{y}_j)$ with a generator $G_{\theta_G}(\mathbf{z}, \mathbf{h}_i, \mathbf{y}_j)$ trained jointly with a discriminator $D_{\theta_D}(\mathbf{x},\mathbf{h}_i, \mathbf{y}_j)$.

\section{Experimental evaluation}

We describe our experimental setup in Section \ref{sec:setup}, followed by results presented in the unlabeled setting in \sect{unsup}, dataset transfer in \sect{transfer} and class-conditional generation  in \sect{supervised}.
We  analyze the impact of the number of stored instances and neighborhood size in 
\sect{ablation}.\looseness-1

\subsection{Experimental setup}
\label{sec:setup}

\textbf{Datasets.} We evaluate our model in the unlabeled scenario 
on  \ImNet~\cite{ILSVRC15} and \COCO~\cite{caesar2018cvpr}. The \ImNet dataset contains 1.2M and 50k images for training and evaluation, respectively. \COCO is a very diverse and complex dataset which contains multi-object images and has been widely used for complex scene generation. 
We use the train and evaluation splits of~\cite{casanova2020generating}, and the (un)seen subsets of the evaluation images with only class combinations that have (not)  been seen during training.
These splits contain 76k, 2k, 675 and 1.3k images, respectively. 
For the class-conditional image generation, we use \ImNet as well as ImageNet-LT~\cite{openlongtailrecognition}.
The latter is a long-tail variant of  \ImNet that contains a subset of 115k samples, where the 1,000 classes have between 5 and 1,280  samples each. Moreover, we use some samples of four additional datasets to highlight the transfer abilities of \ours: Cityscapes~\cite{Cordts2016Cityscapes}, MetFaces~\cite{karras2020training}, PACS~\cite{li2017deeper} and Sketches~\cite{eitz2012hdhso}.

\textbf{Evaluation protocol.} 
We report Fréchet Inception Distance (FID)~\cite{heusel17nips}, Inception Score (IS)~\cite{salimans16nips}, and  LPIPS~\cite{zhang2018perceptual}. 
LPIPS computes the distance between the AlexNet activations of two  images generated with two different latent vectors and same conditioning.
On \ImNet, we follow~\cite{brock2018large}, and compute FID  over 50k generated images and the 50k real validation samples are used as reference.
On \COCO and \ImNet-LT, we compute the FID for each of the splits using all images in the split as reference, and sample the same number images. 
Additionally, in \ImNet-LT we stratify the FID by grouping classes based on the number of train samples:  more than 100  (many-shot FID),  between 20 and 100  (med-shot FID), and  less than 20 (few-shot FID). 
For the reference set, we split the validation images along these three groups of classes, and generate a matching number of samples per group.
In order to compute all above-mentioned metrics, \ours requires instance features for sampling. 
Unless stated otherwise, we store 1,000 training set instances by applying  k-means clustering to the training set and selecting the features of the data point that is the closest to each one of the centroids. 
All quantitative metrics for \ours are reported over five random seeds for the input noise when sampling from the model.

\textbf{Network architectures and hyperparameters.}
As  feature extractor $f_{\phi}$, we use a ResNet50~\cite{he2016deep} trained in a  self-supervised way with SwAV~\cite{caron2020unsupervised} for the unlabeled scenario; for the class-conditional \ours, we use a ResNet50 trained for the classification task on either \ImNet or \ImNet-LT~\cite{kang2019decoupling}. 
For \ImNet experiments, we use BigGAN~\cite{brock2018large} as a baseline architecture, given its superior image quality and ubiquitous use in conditional image generation. For \ours, we replace the class embedding layers in the generator by a fully connected layer that takes the instance features as input and reduces its dimensionality from 2,048 to 512; the same approach is followed to adapt the discriminator. 
For COCO-Stuff, we additionally include the state-of-the-art unconditional StyleGAN2 architecture~\cite{Karras2019stylegan2}, as it has shown good generation quality and diversity in the lower data regime~\cite{karras2020training,Karras2019stylegan2}. We follow its class-conditional version~\cite{karras2020training} to  extend it to \ours by replacing the input class embedding by the instance features. Unless stated otherwise, we set the size of the neighborhoods to $k\!=\!50$ for \ImNet and $k\!=\!5$ for both COCO-Stuff and ImageNet-LT. See the supplementary material for details on the architecture and optimization hyperparameters.

\subsection{Unlabeled setting}
\label{sec:unsup}

\setlength\intextsep{0pt}
\begin{wraptable}{r}{8.7cm}
\caption{Results for  \ImNet in unlabeled setting.
For fair comparison with~\cite{noroozi2020self} at $64\!\times\!64$ resolution, we trained an unconditional BigGAN model and report the non-official FID and IS scores -- computed with Pytorch rather than  TensorFlow -- indicated with *. $^\dagger$: increased parameters to match \ours capacity. DA: 50\% horizontal flips in (\textbf{d}) real and fake samples, and (\textbf{i}) conditioning instances. $ch \times$: Channel multiplier that affects network width as in BigGAN.
\looseness-1}\label{table:unsup_in}
\footnotesize
\centering
\begin{tabular}{@{}llll@{}}
\toprule
\textbf{Method}
&  \textbf{Res.} & $\downarrow$\textbf{FID} & $\uparrow$\textbf{IS} \\  \midrule
Self-sup. GAN ~\cite{noroozi2020self} & 64 & 19.2* & 16.5* \\ %
Uncond.\ BigGAN$^\dagger$ & 64 & 16.9* $\pm$ 0.0 & 14.6* $\pm$ 0.1 \\

\textbf{\ours}  & 64 & 10.4* $\pm$ 0.1 & 21.9* $\pm$ 0.1 \\ 
\textbf{\ours} + DA (\textbf{d},\textbf{i}) & 64 & \textbf{9.2}* $\pm$ 0.0 & \textbf{23.5}* $\pm$ 0.1 \\ 

\midrule

MGAN~\cite{hoang2018mgan} & 128 & 58.9 & 13.2 \\ 
PacGAN2~\cite{lin2018pacgan}& 128 & 57.5 & 13.5 \\
Logo-GAN-AE~\cite{sage2018logo} & 128 & 50.9 & 14.4 \\
Self-cond. GAN~\cite{liu2020diverse}& 128 & 41.7 & 14.9\\
Uncond.\ BigGAN~\cite{pmlr-v97-lucic19a} & 128 & 25.3 & 20.4 \\
SS-cluster GAN~\cite{pmlr-v97-lucic19a}& 128 & 22.0 & 23.5 \\ 
PGMGAN~\cite{armandpour2021partition} & 128 & 21.7 & 23.3 \\ 

\textbf{\ours} & 128 & 13.2 $\pm$ 0.0 & 45.5 $\pm$ 0.2 \\ 
\textbf{\ours} + DA (\textbf{d},\textbf{i}) & 128 & \textbf{11.7} $\pm$ 0.0 & \textbf{48.7} $\pm$ 0.1 \\ 
\midrule 
ADM \cite{dhariwal21arxiv} & 256 &  32.5 & 37.6 \\
\textbf{\ours} ($ch\times64$) & 256 &  17.0 $\pm$ 0.2 & 53.0 $\pm$ 0.4 \\
\textbf{\ours} ($ch\times64$) + DA (\textbf{d},\textbf{i}) & 256 &  17.4 $\pm$ 0.1 & 53.5 $\pm$ 0.5 \\ 
\textbf{\ours} ($ch\times96$) + DA (\textbf{d}) & 256 &  \textbf{15.6} $\pm$ 0.1 & \textbf{59.0} $\pm$ 0.4 \\
\bottomrule
\end{tabular}
\end{wraptable}

\textbf{\ImNet.}
We start by comparing \ours against previous work in Table~\ref{table:unsup_in}. Note that unconditional BigGAN baseline is trained by setting all labels in the training set to zero, following~\cite{pmlr-v97-lucic19a,noroozi2020self}. 
\ours surpasses all previous approaches at both $64\!\times\!64$ and $128\!\times\!128$ resolutions in both FID and IS scores. 
At $256\!\times\!256$ resolution, \ours outperforms the concurrent unconditional diffusion-based model of \cite{dhariwal21arxiv}; the only other result  we are aware of in this setting. Additional results in terms of precision and recall can be found in Table \ref{table:pr_table} in the supplementary material.

As shown in Figure~\ref{subfig:teaser_unsup}, \ours generates high quality images preserving most of the appearance of the conditioning instance. 
Note that generated images are not mere training memorizations; as shown in the supplementary material, generated images differ substantially from the nearest training samples.

\begin{figure}
\centering
\begin{subfigure}[t]{0.098\textwidth}
 \centering
\includegraphics[width=\textwidth]{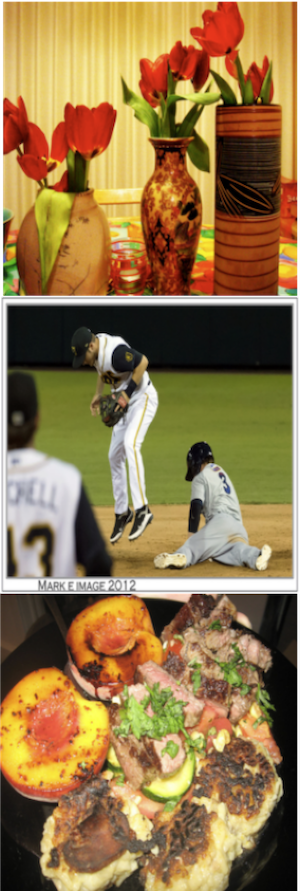}
\caption{$\mathbf{x}_i$}
\label{subfig:coco_sota_gt}
\end{subfigure}
\begin{subfigure}[t]{0.29\textwidth}
 \centering
\includegraphics[width=\textwidth]{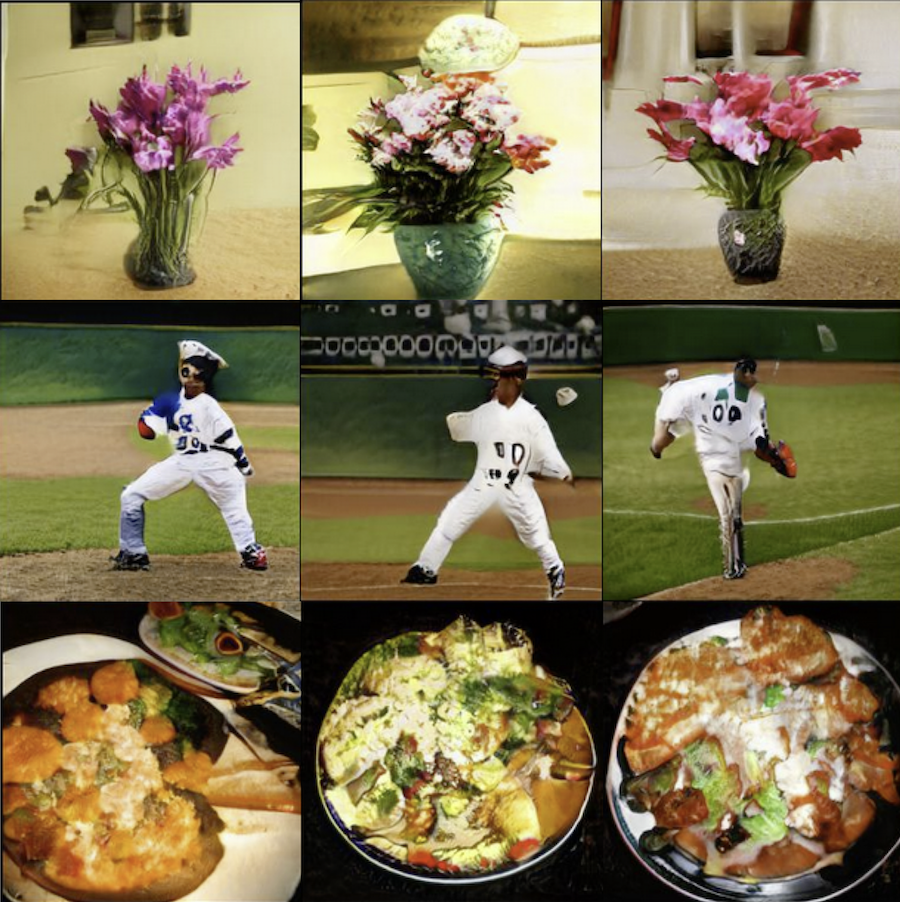}
\caption{\ours (StyleGAN2) }
\label{subfig:coco_qualitative_stylegan}
\end{subfigure}
\begin{subfigure}[t]{0.29\textwidth}
 \centering
\includegraphics[width=\textwidth]{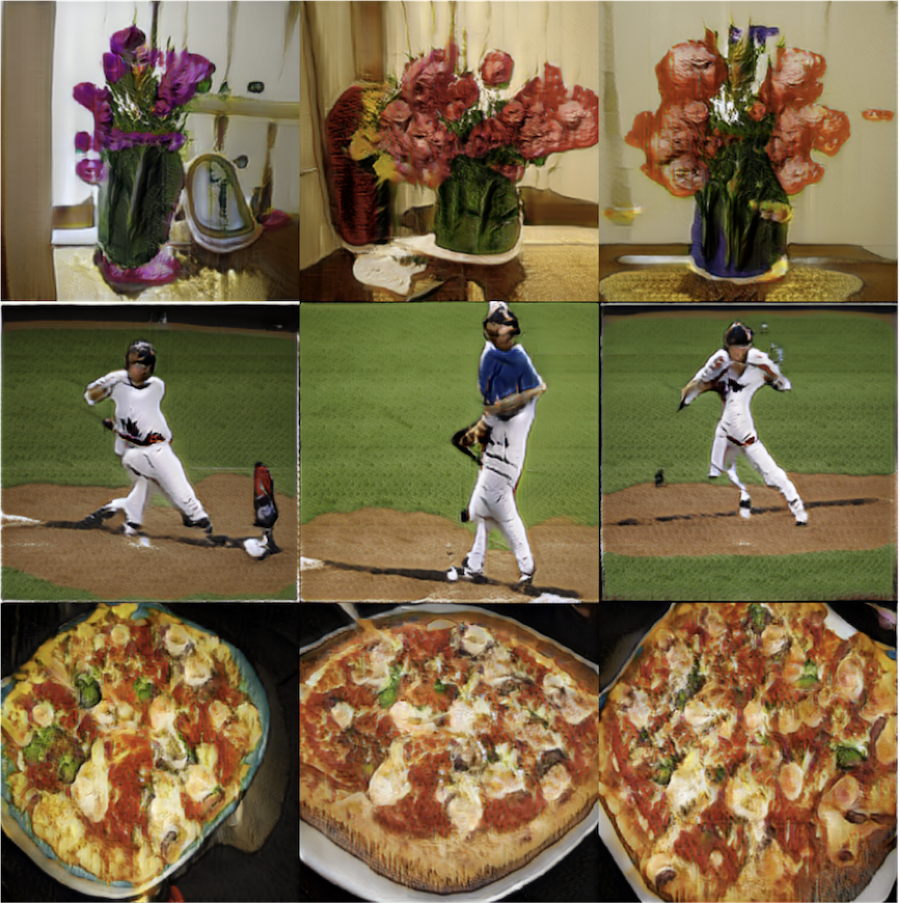}
\caption{\ours (BigGAN) }
\label{subfig:coco_qualitative_biggan}
\end{subfigure}
\begin{subfigure}[t]{0.29\textwidth}
 \centering
\includegraphics[width=\textwidth]{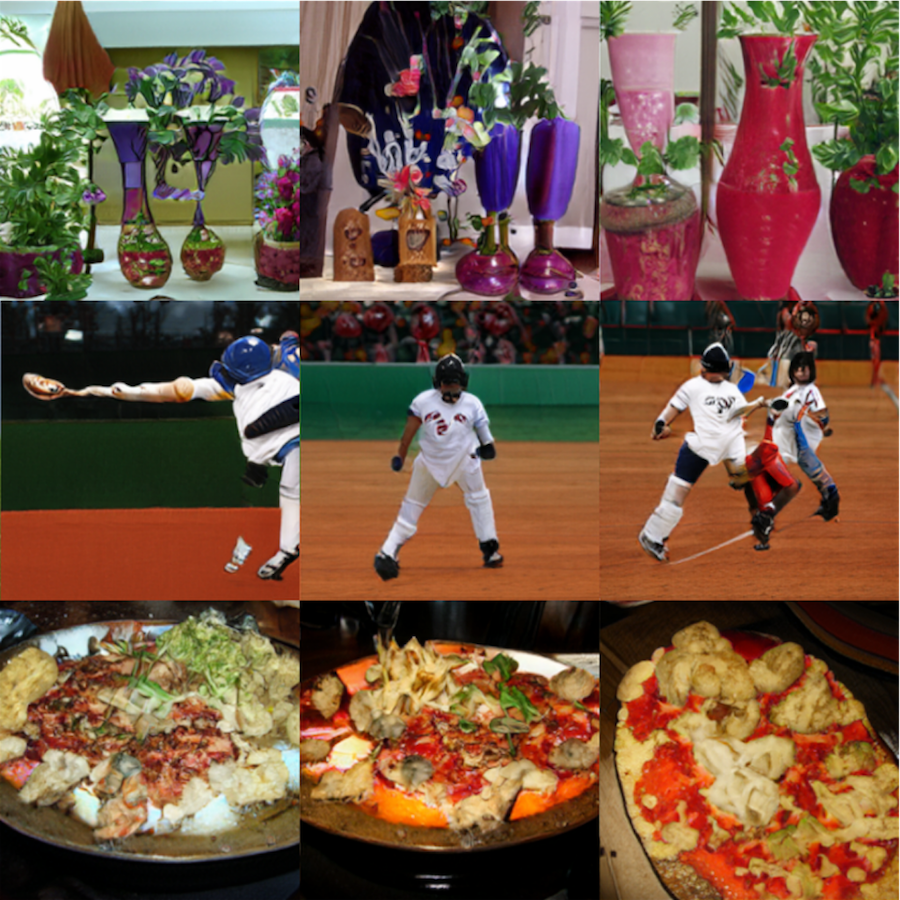}
\caption{\ours (BigGAN, transf.) }
\label{subfig:coco_qualitative_transfer}
\end{subfigure}
\caption{Qualitative comparison for scene generation on $256\!\times\!256$ COCO-Stuff.
}
\label{fig:coco_qualitative}
\end{figure}

\textbf{COCO-Stuff.}
We proceed with the evaluation of \ours on COCO-Stuff in Table~\ref{table:coco_quantitative}. We also compare  to state-of-the-art complex scene generation pipelines which rely on labeled bounding box annotations as conditioning -- LostGANv2~\cite{sun2020learning} and OC-GAN~\cite{sylvain2020object}. Both of these approaches use tailored architectures for complex scene generation, which have at least twice the number of parameters of \ours. 
Our \ours matches or improves upon the unconditional version of the same backbone architecture in terms of FID in all cases, except for training FID with the StyleGAN2 backbone at $256\!\times\!256$ resolution.
Overall, the StyleGAN2 backbone is superior to BigGAN on this dataset, and StyleGAN2-based \ours achieves the state-of-the-art FID scores, even when compared to the bounding-box conditioned LostGANv2 and OC-GAN.
 \ours exhibits notably higher LPIPS than LostGANv2 and OC-GAN, which  could be explained by the fact that  the latter   only leverage one real sample per input conditioning during  training; whereas \ours uses multiple real neighboring samples per each instance, naturally favouring diversity in the generated images. 
 As shown in figures~\ref{subfig:coco_qualitative_stylegan} and~\ref{subfig:coco_qualitative_biggan}, \ours generates high quality diverse images given the input instance.
 A qualitative comparison between LostGANv2, OC-GAN and \ours can be found in Section \ref{app:add_qualitative} of the supplementary material.

 \begin{table}[h]
 \caption{Quantitative results on COCO-Stuff. \ours trained on \ImNet indicated as ``transf''. 
 Some  non-zero standard deviations are reported as  0.0 because of rounding.
}
\label{table:coco_quantitative}
 \footnotesize
\centering
\resizebox{\textwidth}{!}{ 
 \begin{tabular}{@{}lrccccc@{}}
\toprule
  & & \multicolumn{4}{c}{$\downarrow$\textbf{FID}} & $\uparrow$ \textbf{LPIPS}\\
 128$\times$128 & \rotatebox[origin=c]{0}{\textbf{\# prms.}}  &  \rotatebox[origin=c]{0}{\textbf{train}}  &  \rotatebox[origin=c]{0}{\textbf{eval}}   &
 \rotatebox[origin=c]{0}{\textbf{eval seen}}
&  \rotatebox[origin=c]{0}{\textbf{eval unseen}} & \rotatebox[origin=c]{0}{\textbf{eval}}  \\  \midrule
LostGANv2~\cite{sun2020learning} &  41 M &  12.8 $\pm$ 0.1 & 40.7 $\pm$ 0.3 & 80.0 $\pm$ 0.4 & 55.2 $\pm$ 0.5 & 0.45 $\pm$ 0.1 \\
OC-GAN~\cite{sylvain2020object} &  170 M & --- & 45.1 $\pm$ 0.3 & 85.8 $\pm$ 0.5 & 60.1 $\pm$ 0.2 & 0.13 $\pm$ 0.1 \\
Unconditional (BigGAN) & 18 M & 17.9 $\pm$ 0.1 & 46.9 $\pm$ 0.5 & 103.8 $\pm$ 0.8 & 60.9 $\pm$ 0.7 & 0.68 $\pm$ 0.1\\
\ours (BigGAN) &  22 M & 16.8 $\pm$ 0.1 & 44.9 $\pm$ 0.5 & 81.5 $\pm$ 1.3 & 60.5 $\pm$ 0.5 & 0.67 $\pm$ 0.1 \\
\ours   (BigGAN, transf.) &  77 M & \textbf{8.5} $\pm$ 0.0 & \textbf{35.6} $\pm$ 0.2 & 77.0 $\pm$ 1.0 & \textbf{48.9} $\pm$ 0.2 & \textbf{0.69} $\pm$ 0.1\\
Unconditional (StyleGAN2) &  23 M & 8.8 $\pm$ 0.1 & 37.8 $\pm$ 0.2 & 92.1 $\pm$ 1.0 & 53.2 $\pm$ 0.5 & 0.68 $\pm$ 0.1 \\
\ours (StyleGAN2) &  24 M & 8.9 $\pm$ 0.0 & 36.2 $\pm$ 0.2 & \textbf{74.3} $\pm$ 0.8 & 50.8 $\pm$ 0.3 &  0.67 $\pm$ 0.1 \\
\midrule
256$\times$256 \\ \midrule
LostGANv2~\cite{sun2020learning} &  46 M & 18.0 $\pm$ 0.1 & 47.6 $\pm$ 0.4 & 88.5 $\pm$ 0.4 & 62.0 $\pm$ 0.6 & 0.56 $\pm$ 0.1\\
OC-GAN~\cite{sylvain2020object} &  190 M & --- & 57.0 $\pm$ 0.1 & 98.7 $\pm$ 1.2 & 71.4 $\pm$ 0.5 & 0.21 $\pm$ 0.1\\
Unconditional (BigGAN) & 21 M & 51.0 $\pm$ 0.1 & 81.6 $\pm$ 0.5 & 135.1 $\pm$ 1.6  & 95.8 $\pm$ 1.1 & \textbf{0.77} $\pm$ 0.1 \\
\ours  (BigGAN) & 26 M & 24.6 $\pm$ 0.1 & 53.1 $\pm$ 0.4 & 88.5 $\pm$ 1.8 & 69.1 $\pm$ 0.6 & 0.73 $\pm$ 0.1\\
\ours  (BigGAN, transf.) &  90 M & 13.9 $\pm$ 0.1 & \textbf{40.9} $\pm$ 0.3 & 79.4 $\pm$ 1.2 & \textbf{55.6} $\pm$ 0.6 & 0.76 $\pm$ 0.1  \\
Unconditional (StyleGAN2) & 23 M & \textbf{7.1} $\pm$ 0.0 & 44.6 $\pm$ 0.4 & 98.1 $\pm$ 1.7 & 59.9 $\pm$ 0.5 & 0.76 $\pm$ 0.1 \\
\ours(StyleGAN2) & 25 M & 9.6 $\pm$ 0.0 & 41.4 $\pm$ 0.2 & \textbf{76.7} $\pm$ 0.6 & 57.5 $\pm$ 0.5 & 0.74 $\pm$ 0.1  \\
\bottomrule
\end{tabular}
}
\end{table}

\subsection{Off-the-shelf transfer to other datasets}
\label{sec:transfer}
In our first transfer experiment, we train \ours  with a BigGAN architecture on \ImNet, and use it to generate images from COCO-Stuff instances at  test time. Quantitative results are reported as ``\ours (transf.)'' in Table~\ref{table:coco_quantitative}. 
In this setup, no \COCO images  are used to train  the model, thus, all splits contain unseen objects combinations. 
Perhaps surprisingly, \ours trained on \ImNet  outperforms the same model trained on COCO-Stuff for all splits: 8.5 \vs 16.8 train FID at 128 resolution. This  raises the question of how close \ImNet and COCO-Stuff data distributions are. 
We compute the FID between real data train split of the two datasets at $128\!\times\!128$ resolution and obtain a score of $37.2$. Hence, the remarkable transfer capabilities of \ours are not 
explained by  dataset similarity and
may be attributed to the effectiveness of the \ImNet pre-trained  feature extractor and generator.
When we replace the conditioning instances from \COCO with those of \ImNet, we obtain a train FID score of 43.5, underlining the important distribution shift that can be implemented by changing the conditioning instances. 

Interestingly, the transferred \ours also outperforms LostGANv2 and OC-GAN  which condition on  labeled bounding box annotations.
Transferring the model from \ImNet  boosts diversity \wrt the model trained on COCO-Stuff (see LPIPS in Table \ref{table:coco_quantitative}), which may be in part due to the   larger $k\!=\!50$ used for \ImNet training, compared to $k\!=\!5$ when training on \COCO.  Qualitative results of COCO-Stuff generations from the \ImNet pre-trained \ours 
can be found in Figure~\ref{subfig:teaser_unsup_transfer} (top row) and Figure~\ref{subfig:coco_qualitative_transfer}.
These generations suggest that \ours is able to effectively leverage the large scale training on 
\ImNet to improve the quality and diversity of the COCO-Stuff scene generation, which contains significantly less data to train. 
 
We further explore how  the \ImNet trained \ours transfers to  conditioning on other datasets using Cityscapes, MetFaces, and PACS  in Figure~\ref{subfig:teaser_unsup_transfer}. 
Generated images still preserve the semantics and style of the images for all datasets, although degrading their quality when compared to samples in Figure~\ref{subfig:teaser_unsup}, as the instances in these datasets --in particular MetFaces and PACS-- are very different from the \ImNet ones.
See Section \ref{app:add_transfer} in the supplementary material for more discussion,  additional evaluations, and more qualitative examples of dataset transfer.

\subsection{Class-conditional setting}
\label{sec:supervised}
\setlength\intextsep{0pt}
    \begin{wraptable}{r}{9cm}
    \caption{Class-conditional  results on \ImNet.
    \mbox{*: Trained} using open source  code. DA: 50\% horizontal flips in (\textbf{d}) real and fake samples, and (\textbf{i}) conditioning instances. $ch \times$: Channel multiplier that affects network width. $^\dagger$: numbers from the original paper, as training diverged with the BigGAN opensourced code.
    }\label{tab:quantitative_class_conditional}
    \small
    \centering
     \begin{tabular}{@{}lcrr@{}}
    \toprule
     & \textbf{Res.} &  $\downarrow$\textbf{FID} & $\uparrow$\textbf{IS} \\ \midrule 
    BigGAN*~\cite{brock2018large} & 64  & 12.3 $\pm$ 0.0 & 27.0 $\pm$ 0.2 \\ 
    BigGAN*~\cite{brock2018large} + DA (\textbf{d}) & 64  & 10.2 $\pm$ 0.1 & 30.1 $\pm$ 0.1 \\ 
     \ours & 64 & 8.5 $\pm$ 0.0 & 39.7 $\pm$ 0.2 \\ 
     \ours + DA(\textbf{d}, \textbf{i}) & 64 & \textbf{6.7} $\pm$ 0.0 & \textbf{45.9} $\pm$ 0.3 \\
    \midrule
    BigGAN*~\cite{brock2018large} & 128 & 9.4 $\pm$ 0.0 & 98.7 $\pm$ 1.1 \\ 
    BigGAN*~\cite{brock2018large} +  DA(\textbf{d}) & 128 & \textbf{8.0} $\pm$ 0.0 & 107.2 $\pm$ 0.9 \\
     \ours & 128 & 10.6 $\pm$ 0.1 & 100.1 $\pm$ 0.5 \\ 
     \ours + DA(\textbf{d}, \textbf{i}) & 128 & 9.5 $\pm$ 0.1 & \textbf{108.6} $\pm$ 0.7 \\ 
    \midrule
    BigGAN*~\cite{brock2018large} ($ch\times64$) & 256 & 8.0 $\pm$ 0.1 &  139.1 $\pm$ 0.3 \\ 
    BigGAN*~\cite{brock2018large} ($ch\times64$) + DA(\textbf{d})& 256 & 8.3 $\pm$ 0.1 &  125.0 $\pm$ 1.1 \\
     \ours ($ch\times64$) & 256 & 8.3 $\pm$ 0.1 & 143.7 $\pm$ 1.1 \\ 
     \ours ($ch\times64$) + DA(\textbf{d}, \textbf{i}) & 256 & \textbf{7.5} $\pm$ 0.0 & 152.6 $\pm$ 1.1 \\
      BigGAN$^\dagger$~\cite{brock2018large} ($ch\times96$) & 256 & 8.1 & 144.2 \\
    \ours ($ch\times96$) + DA(\textbf{d}) & 256 & 8.2 $\pm$ 0.1 & \textbf{173.8} $\pm$ 0.9 \\
    \bottomrule
    \end{tabular}
    \end{wraptable}
    
\textbf{\ImNet.} In Table~\ref{tab:quantitative_class_conditional}, we show that the class-conditioned \ours outperforms BigGAN in terms of both FID and IS across all resolutions except the FID at $128\!\times\!128$ resolution. 
It is worth mentioning that, unlike BigGAN, \ours can control the semantics of the generated images by either fixing the instance features and swapping the class conditioning, or by fixing the class conditioning and swapping the instance features;
see Figure~\ref{subfig:teaser_class}. As shown in the figure, generated images preserve  semantics of both the class label and the instance, generating different dog breeds on similar backgrounds, or generating camels in the snow, an unseen scenario in \ImNet to the best of our knowledge. Moreover, in Figure~\ref{subfig:teaser_class_transfer}, we show the transfer capabilities of our class-conditional \ours trained on \ImNet and conditioned on instances from other datasets, generating camels in the grass, zebras in the city, and husky dogs with the style of MetFaces and PACS instances. 
These controllable conditionings enable the generation of images that are not present or very rare in the \ImNet dataset, \eg camels surrounded by snow or zebras in the city.
Additional qualitative transfer results which either fix the class label and swap the instance features, or vice-versa,  can be found in 
Section \ref{app:add_transfer} of
the supplementary material.

\textbf{ImageNet-LT.}  
Due to the class imbalance in \ImNet-LT, selecting a subset of instances with either k-means or uniform sampling can easily result in ignoring rare classes,  and penalizing their generation. 
Therefore, for this dataset we use all available 115k training instances to sample from the model and compute the metrics. In Table~\ref{table:imagenet_lt_quantitative} we compare to BigGAN, showing that \ours is better in terms of FID and IS for modeling this long-tailed distribution.  
Note that the improvement is noticeable for each of the three groups of classes with different number of samples, see  many/med/few column.
In Section \ref{app:class_balacing} of the supplementary material 
we present experiments when using  class-balancing to train BigGAN, showing that it does not improve quality nor diversity of generated samples. We hypothesize that oversampling some classes may result in overfitting for the discriminator, leading to low quality image generations.

\subsection{Selection of stored instances and neighborhood size}
\label{sec:ablation}

In this section, we empirically justify the k-means procedure to select the instances to sample from the model,  consider the effect of the number of instances used to sample from the model, as well as the effect of the size $k$ of the neighborhoods $\mathcal{A}_i$ used during training. 
The impact of different choices for the instance embedding function $f_\phi(\mathbf{x})$ is evaluated in the supplementary material.

\begin{table}[t]
\footnotesize
\centering
\caption{ Class-conditional results on ImageNet-LT. 
\mbox{*: Trained} using open source  code.}
{
 \begin{tabular}{@{}lcccccc@{}}
\toprule
 & \textbf{Res.} & $\downarrow$\textbf{train FID} & $\uparrow$\textbf{train IS} & $\downarrow$\textbf{val FID} & \textbf{many/med/few $\downarrow$val FID } & $\uparrow$\textbf{val IS} \\  \midrule
BigGAN*~\cite{brock2018large} & 64 & 27.6 $\pm$ 0.1 & 18.1 $\pm$ 0.2 & 28.1 $\pm$ 0.1  & 28.8 / 32.8 / 48.4 $\pm$ 0.2 & 16.0 $\pm$ 0.1  \\
 \ours & 64 & \textbf{23.2} $\pm$ 0.1 & \textbf{19.5} $\pm$ 0.1 & \textbf{23.4} $\pm$ 0.1 & \textbf{23.8 / 28.0 / 42.7} $\pm$ 0.1 & \textbf{17.6}  $\pm$ 0.1 \\
\midrule
BigGAN*~\cite{brock2018large} & 128 & 31.4 $\pm$ 0.1 & 30.6 $\pm$ 0.1 & 35.4 $\pm$ 0.1 & 34.0 / 43.5 / 64.4 $\pm$ 0.2 & 24.9 $\pm$ 0.2 \\
 \ours & 128 &  \textbf{23.4} $\pm$ 0.1 & \textbf{39.6} $\pm$ 0.2 & \textbf{24.9} $\pm$ 0.1 & \textbf{24.3 / 31.4 / 53.6} $\pm$ 0.3 & \textbf{32.5} $\pm$ 0.1 \\
\midrule
BigGAN*~\cite{brock2018large} & 256 & 27.8 $\pm$ 0.0 & 58.2 $\pm$ 0.2 & 31.4 $\pm$ 0.1 & 28.1 / 40.9 / 67.6 $\pm$ 0.3 & 44.7 $\pm$ 0.2 \\
 \ours & 256 & \textbf{21.7} $\pm$ 0.1 & \textbf{66.5} $\pm$ 0.3 & \textbf{23.4} $\pm$ 0.1 & \textbf{20.6 / 32.4 / 60.0}  $\pm$ 0.2 & \textbf{51.7} $\pm$ 0.1 \\
\bottomrule
\end{tabular}
}
 \label{table:imagenet_lt_quantitative}
 \end{table}

\begin{figure}[b]
\centering
\begin{subfigure}{0.49\textwidth}
 \centering
\includegraphics[width=\textwidth]{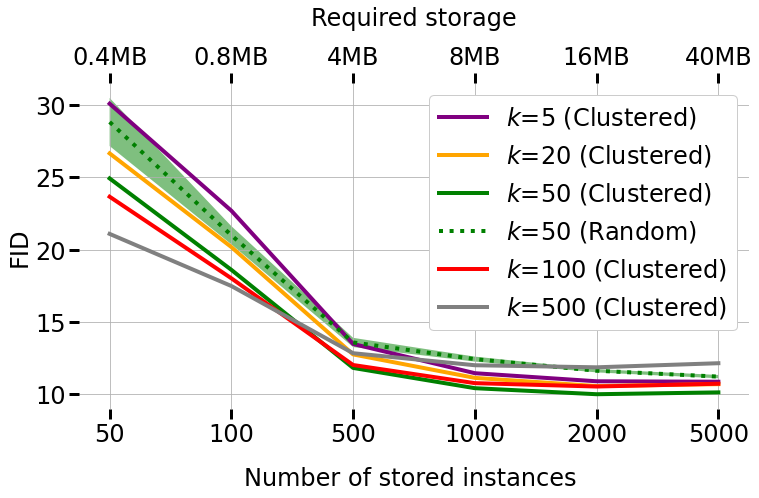}
\label{subfig:ablation_instances1}
\end{subfigure}
\begin{subfigure}{0.49\textwidth}
 \centering
\includegraphics[width=\textwidth]{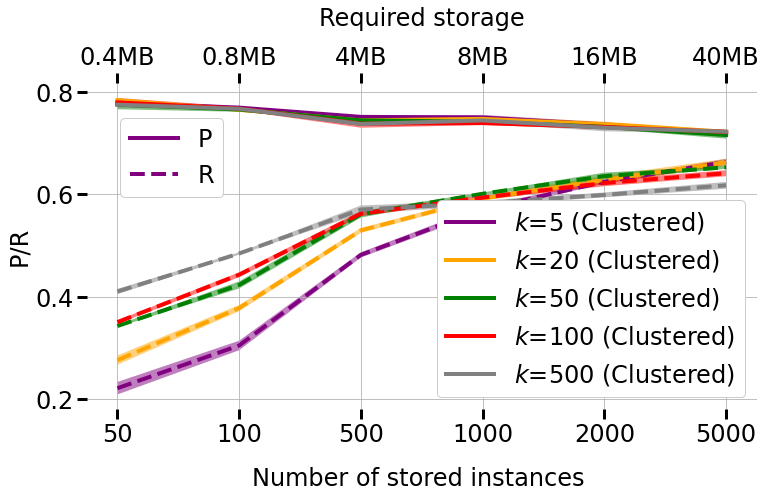}
\label{subfig:ablation_instances2}
\end{subfigure}

\caption{Impact on the number of stored instances used to evaluate \ours and the size of the neighborhood $k$.
Experiments performed on the $64\!\times\!64$ unlabeled \ImNet dataset.
}
\label{fig:ablation_instances_main}
\vspace{-0.0cm}
\end{figure}

\textbf{Selecting instances to sample from the model.}
\label{subsubsec:instance_selection}
In Figure~\ref{fig:ablation_instances_main} (left), we compare two instance selection methods in terms of FID: uniform sampling (Random) and k-means (Clustered), where we select the closest instance to each cluster centroid,  using  $k\!=\!50$ neighbors during training (solid and dotted green  lines). 
Random selection is  consistently outperformed by k-means; selecting only 1,000 instances with k-means results in better FID than randomly selecting 5,000 instances. Moreover, storing more than 1,000 instances selected with k-means does not result in noticeable improvements in  FID. 
Additionally, we computed FID metrics for the 1,000 ground truth images that are closest to the k-means cluster centers,
obtaining $41.8 \pm 0.2$ FID, which is considerably higher than the $10.4 \pm 0.1 $ FID we obtain with \ours ($k=50$) when using the same
1,000 cluster centers.
This supports the idea that \ours is generating data points that go beyond the stored instances, better recovering the data distribution. 

We consider precision (P) and recall (R)~\cite{kynkaanniemi2019improved} (using an InceptionV3~\cite{DBLP:journals/corr/SzegedyVISW15} as feature extractor and sampling 10,000 generated and real images) to disentangle the factors driving the improvement in FID, namely image quality and diversity (coverage) -- see Figure \ref{fig:ablation_instances_main} (right). We see that augmenting the number of stored instances results in slightly worse precision (image quality) but notably better recall (coverage). Intuitively, this suggests that by increasing the number of stored instances, we can better recover the data density at the expense of slightly degraded image quality in lower density regions of the manifold -- see \eg \cite{devries2020instance}.\looseness-1

\textbf{Neighborhood size.}
In Figure~\ref{fig:ablation_instances_main} (both panels) we analyze the interplay between the neighborhood size and the number of instances used to recover the data distribution. For small numbers of stored instances, 
we observe that  larger the neighborhoods lead to better (lower)  FID scores (left-hand side of left panel).  
For recall, we also observe improvements for large neighborhoods when storing few instances  (left-hand side of right panel), suggesting that larger neighborhoods are more effective in recovering the data distribution from few instances. 
This trend is reverted for large numbers of stored instances, where smaller values of \emph{k} are more effective. 
This supports the idea that the neighborhood size acts as a bandwidth parameter -- similar to KDE --, that controls the smoothness of the implicitly learnt conditional distributions around instances. For example, $k=500$ leads to smoother conditional distributions than $k=5$, and as a result requires fewer stored instances to recover the data distribution. Moreover, as expected, we notice that the value of $k$ does not significantly affect precision (right panel).  Overall, $k=50$ offers a good compromise, exhibiting top performance across all metrics when using at least $500$ stored instances.
We visualize the smoothness effect by means of a qualitative comparison across samples from different neighborhood sizes in Section \ref{app:k_size} of the supplementary material. Using (very) small neighborhoods (\eg of $k=5$), results in lower diversity in the generated images. \looseness-1

\section{Related work}
\textbf{Data partitioning for GANs.}
Previous works have attempted to improve the image generation quality and diversity of GANs 
by partitioning the data manifold through clustering techniques~\cite{armandpour2021partition, grinblat2017class, liu2020diverse, pmlr-v97-lucic19a, noroozi2020self, sage2018logo}, or by leveraging mixture models in their design~\cite{eghbal2019mixture,Ghosh_2018_CVPR,hoang2018mgan}. 
In particular, \cite{pmlr-v97-lucic19a,sage2018logo} apply k-means on representations from a pre-trained feature extractor to cluster the data,  and then use cluster indices to condition the generator network. Then, \cite{grinblat2017class,liu2020diverse} introduce an alternating two-stage approach where the first stage applies k-means to the discriminator feature space and the second stage trains a GAN conditioned on the cluster indices. 
Similarly,~\cite{noroozi2020self} proposes to train a clustering network, which outputs pseudolabels, in cooperation with the generator. 
Further,~\cite{armandpour2021partition} trains a feature extractor with self-supervised pre-training tasks, and creates a k-nearest neighbor graph in the learned representation space to cluster connected points into the same sub-manifold. In this case, a different generator is then trained for each identified sub-manifold. 
By contrast, \ours uses fine-grained overlapping data neighborhoods in tandem with conditioning on rich feature embeddings (instances) to learn a localized distribution around each data point.\looseness-1

\textbf{Mitigating mode collapse in GANs.} Works which attempt to mitigate mode collapse may also bear some similarities to ours. 
In~\cite{lin2018pacgan}, the discriminator takes into consideration multiple random samples from the same class to output a decision. In~\cite{lucas18icml}, a mixed batch of generated and real samples is fed to the discriminator with the goal of predicting the ratio of real samples in the batch. Other works use a mixture of generators~\cite{Ghosh_2018_CVPR,hoang2018mgan} and encourage each generator to focus on generating samples from a different mode. Similarly, in~\cite{eghbal2019mixture}, the discriminator is pushed to form clusters in its representation space, where each cluster is represented by a Gaussian kernel. In turn, the generator tends to learn to generate samples covering all clusters, hence mitigating mode collapse. By contrast, we focus on discriminating between real and generated \emph{neighbors} of an instance conditioning, by using a single generator network trained following the GAN formulation.

\textbf{Conditioning on feature vectors}. Very recent work~\cite{mangla2021data} uses image self-supervised feature representations to condition a generative model whose objective is to produce a good input reconstruction; this requires storing the features of all training samples. In contrast, our objective is to learn a localized distribution (as captured by nearest neighboring images) around each conditioning instance, and we only need to save a very small subset of the dataset features to approximately recover the training distribution.

\textbf{Kernel density estimation and adversarial training.}  Connections between adversarial training and nonparametric density estimation   have been made in  prior work~\cite{ abbasnejad2019generative}.
However, to the best of our knowledge, no prior work models  the dataset density  in a nonparametric fashion with a localized distribution around each data point with a single conditional  generation network.

\textbf{Complex scene generation.}  Existing methods for complex scene generation, where natural looking scenes contain multiple objects, most often aim at controllability and rely on detailed conditionings such as a scene graphs~\cite{ashual2019specifying,johnson2018image}, bounding box layouts~\cite{sun2019image,sun2020learning,sylvain2020object,Zhao_2019_CVPR}, semantic segmentation masks~\cite{chen2017photographic,  park2019SPADE,qi2018semi, tang2020local,wang2018high} or more recently, freehand sketches~\cite{gao2020sketchycoco}. All these methods leverage intricate pipelines to generate complex scenes and require labeled datasets. By contrast, our approach relies on instance conditionings which control the global semantics of the generation process, and does not require any dataset labels.
It is worth noting that complex scene generation is often characterized by unbalanced, strongly long tailed datasets. Long-tail class distributions negatively affect class-conditional GANs, as they struggle to generate visually appealing samples for classes in the tail~\cite{casanova2020generating}. 
However, to the best of our knowledge, no other previous work tackles this problem for GANs.\looseness-1

\section{Discussion}
\label{sec:discussion}

\textbf{Contributions.}
We presented instance-conditioned GAN (\ours), which  models  dataset distributions in a non-parametric way by conditioning both  generator and discriminator on instance features. 
We validated our approach on the unlabeled setting, showing consistent improvements over baselines on \ImNet and \COCO. 
Moreover, we showed through transfer experiments, where we condition the \ImNet-trained model on instances of other datasets, the ability of \ours  to produce compelling samples from different data distributions. 
Finally, we validated \ours in the class-conditional setting, obtaining competitive results on \ImNet and surpassing the BigGAN baseline on the challenging  \ImNet-LT; and showed compelling controllable generations by swapping the class-conditioning given a fixed instance or the instance given a fixed conditioning.

\textbf{Limitations.}
\ours showed excellent image quality for labeled (class-conditional) and unlabeled image generation. However, as any machine learning tool, it has some limitations. 
First, as kernel density estimator approaches, \ours requires storing training instances to use the model.
Experimentally, we noticed that for complex datasets, such as \ImNet, using 1,000 instances is enough to approximately cover the dataset distribution. 
Second, the instance feature vectors used to condition the model are obtained with a pre-trained feature extractor (self-supervised in the unlabeled case) and depend on it.
We speculate that this limitation might be mitigated if the feature extractor and the generator are trained jointly, and leave it as future work. 
Third, although, we highlighted excellent transfer potential of our approach to unseen datasets, we observed that, in the case of transfer to datasets that are \emph{very} different from \ImNet, the quality of generated images degrades. 

\textbf{Broader impacts.}
\ours brings with it several benefits such as excellent image quality in labeled (class-conditional) and unlabeled image generation tasks, and the transfer potential to unseen datasets, enabling the use of our model on a variety of datasets without the need of fine-tuning or re-training. Moreover, in the case of class-conditional image generation, \ours enables controllable generation of content by adapting either the style -- by changing the instance -- or the semantics -- by altering the class --. Thus, we expect that our model can positively affect the workflow for creative content generators. 
That being said, with improving image quality in generative modeling, there is some potential for misuse. A common example are \emph{deepfakes}, where a generative model is used to manipulate images or videos well enough that humans cannot distinguish real from fake, with the intent to misinform. 
We believe, however, that open research on generative image models also contributes to better  understand such synthetic content, and to detect it where it is undesirable. 
Recently, the community has also started to undertake explicit efforts towards detecting  manipulated content by organizing challenges such as the Deepfake Detection Challenge \cite{deepfakes}.

{\small
\bibliographystyle{plainnat}
\bibliography{biblio}
}

\newpage

\normalsize
\begin{center}
\Large{\textbf{Instance-Conditioned GAN: Supplementary Material}}
\end{center}
\appendix
We provide additional material to support the main paper. We credit the used assets by citing their web links and licenses in Section~\ref{app:assets}, and continue by describing the experimental setup and used hyperparameters in Section~\ref{app:hyperparam}. We compute Precision and Recall metrics on \ImNet in Section~\ref{app:pr_metrics}, and we further compare BigGAN and StyleGAN2 backbones for IC-GAN on \ImNet in Section~\ref{app:stylegan_biggan}. We provide additional qualitative results for both \ours on \ImNet in Section~\ref{app:add_qualitative} and \ours off-the-shelf transfer results on other datasets in Section~\ref{app:add_transfer}. Moreover, we provide results when training BigGAN with class balancing on \ImNet-LT in Section~\ref{app:class_balacing}. Finally, we show further impact studies such as the choice of feature extractor (Section~\ref{app:feature_extractor}), the number of conditionings used during training (Section~\ref{app:num_cond_train}), matching storage requirements for unconditional counterparts of BigGAN and StyleGAN2 and \ours (Section~\ref{app:fair_comparison}) and the qualitative impact of neighborhood size $k$ for \ImNet, as well as quantitative results for ImageNet-LT and COCO-Stuff (Section~\ref{app:k_size}).

\section{Assets and licensing information}
\label{app:assets}
In Tables~\ref{table:assets} and~\ref{table:licenses}, we provide the links to the used datasets, repositories and their licenses. We use Faiss~\cite{JDH17} for a fast computation of nearest neighbors and k-means algorithm leveraging GPUs, DiffAugment~\cite{zhao2020differentiable} for additional data augmentation when training BigGAN, and the pre-trained SwAV~\cite{caron2020unsupervised} and ResNet50 models on \ImNet-LT~\cite{kang2019decoupling} to extract instance features. 

\begin{table}[h]
\centering
\small
\caption{Links to the assets used in the paper.}
\begin{tabular}{@{}lc@{}}
\toprule
 Name & GitHub link \\
  \midrule
  \ImNet~\cite{ILSVRC15} & \url{https://www.image-net.org} \\ 
  \ImNet-LT~\cite{openlongtailrecognition} & \url{https://github.com/zhmiao/OpenLongTailRecognition-OLTR}\\ 
COCO-Stuff~\cite{caesar2018cvpr} & \url{https://cocodataset.org/} \\ 
Cityscapes~\cite{Cordts2016Cityscapes} & \url{https://www.cityscapes-dataset.com/} \\ 
MetFaces~\cite{karras2020training} & \url{https://github.com/NVlabs/metfaces-dataset}\\
PACS~\cite{li2017deeper} & \url{https://domaingeneralization.github.io/}\\ 
Sketches~\cite{eitz2012hdhso} & \url{http://cybertron.cg.tu-berlin.de/eitz/projects/classifysketch/}\\ 
\midrule
BigGAN~\cite{brock2018large} & \url{https://github.com/ajbrock/BigGAN-PyTorch} \\
StyleGAN2~\cite{karras2020training} & \url{https://github.com/NVlabs/stylegan2-ada-pytorch}\\
Faiss~\cite{JDH17} & \url{https://github.com/facebookresearch/faiss} \\
DiffAugment~\cite{zhao2020differentiable} & \url{https://github.com/mit-han-lab/data-efficient-gans} \\
PRDC~\cite{naeem2020reliable} & \url{https://github.com/clovaai/generative-evaluation-prdc} \\
SwAV~\cite{caron2020unsupervised} & \url{https://github.com/facebookresearch/swav} \\
Pre-trained ResNet50~\cite{kang2019decoupling} & \url{https://github.com/facebookresearch/classifier-balancing} \\
\bottomrule
\end{tabular}
 \label{table:assets}
 \end{table}

 \begin{table}[h]
 \centering
 \small
 \caption{Assets licensing information.}
\begin{tabular}{@{}lc@{}}
\toprule
 Name & License \\
  \midrule
\ImNet~\cite{ILSVRC15} and \ImNet-LT~\cite{openlongtailrecognition} & Terms of access: \url{https://www.image-net.org/download.php} \\ 
COCO-Stuff~\cite{caesar2018cvpr} & \url{https://www.flickr.com/creativecommons} \\ 
Cityscapes~\cite{Cordts2016Cityscapes} & \url{https://www.cityscapes-dataset.com/license} \\ 
MetFaces~\cite{karras2020training} & Creative Commons BY-NC 2.0 \\
PACS~\cite{li2017deeper} & Unknown \\ 
Sketches~\cite{eitz2012hdhso} & Creative Commons Attribution 4.0 International \\ 
\midrule
BigGAN~\cite{brock2018large} & MIT \\
StyleGAN2~\cite{karras2020training}  & NVIDIA Source Code \\
Faiss~\cite{JDH17}  & MIT \\
DiffAugment~\cite{zhao2020differentiable} & BSD 2-Clause "Simplified" \\
PRDC~\cite{naeem2020reliable} & MIT \\
swAV~\cite{caron2020unsupervised}  & Attribution-NonCommercial 4.0 International \\
Pre-trained ResNet50~\cite{kang2019decoupling} & BSD \\
\bottomrule
\end{tabular}
\label{table:licenses}
 \end{table}

\section{Experimental setup and hyperparameters}
\label{app:hyperparam}
We divide the experimental section into architecture modifications in Subsection~\ref{subsec:mod} and training and hyperparameter details in Subsection~\ref{subsec:tr_details}.
\subsection{Architecture modifications for \ours.}
\label{subsec:mod}
In our \ours experiments, we leveraged BigGAN and  StyleGAN2 backbones, and extended their architectures to handle the introduced instance conditionings. 

When using BigGAN as a base architecture, \ours replaces the class embedding layers in both generator and discriminator by fully connected layers. The fully connected layer in the generator has an input size of $2,\!048$ (corresponding to the feature extractor $f_\theta$'s dimensionality) and an output size $o_{dim}$ that can be adjusted. For all our experiments, we used $o_{dim}\,=\,512$ -- selected out of $\{256, 512, 1,\!024, 2,\!048\}$. The fully connected layer in the discriminator has a variable output size $n_{dim}$ to match the dimensionality of the intermediate unconditional discriminator feature vector -- following the practice in BigGAN~\cite{brock2018large}. 
For the class-conditional \ours, we use both the class embedding layers as well as the fully connected layers associated with the instance conditioning. In particular, we concatenate class embeddings (of dimensionality $c_{dim}\,=\,128$) and instance embeddings (with dimensionality $o_{dim}\,=\,512$). To avoid the rapid growth of parameters when using both class and instance embeddings, we use $n_{dim}/2$ as the output dimensionality for each of the embeddings in the discriminator, so that the resulting concatenation has a dimensionality of $n_{dim}$.

When using StyleGAN2 as a base architecture, we modify the class-conditional architecture of~\cite{karras2020training}. In particular, we replace the class embeddings layers with a fully connected layer of output dimensionality $512$ in the generator. The fully connected layer substituting the class embedding in the discriminator is of variable size. In this case, the instance features are concatenated with the noise vector at the input of the StyleGAN2's mapping network, creating a \emph{style vector} for the generator. However, when it comes to the discriminator, the mapping network is only fed with the extracted instance features to obtain a modulating vector that is multiplied by the internal discriminator representation at each block.

All instance feature vectors $\mathbf{h}_i$ are normalized with $\ell_2$ norm before computing the neighborhoods and when used as conditioning for the GAN.

\subsection{Training details and hyperparameters}
\label{subsec:tr_details}
All models were trained while monitoring the training FID, and training was stopped according to either one of the following criteria: (1) early stopping when FID did not improve for $50$ epochs -- or the equivalent number of iterations depending on the batch size --, or (2) when the training FID diverged. For BigGAN, we mainly explored the hyperparameter space around previously known and successful configurations~\cite{brock2018large,noroozi2020self}. Concretely, we focused on finding the following best hyperparameters for each dataset and resolution: the batch size ($BS$), model capacity controlled by channel multipliers ($CH$), number of discriminator updates versus generator updates ($D_{updates}$), discriminator learning rate ($D_{lr}$) and generator learning rate ($G_{lr}$), while keeping all other parameters unchanged~\cite{brock2018large}. For StyleGAN, we also performed a hyperparameter search around previously known successful settings~\cite{karras2020training}. More precisely, we searched for the optimal $D_{lr}$ and $G_{lr}$ and R1 regularization weight $\gamma$ and used default values for the other hyperparameters.

\paragraph{\ImNet.} When using the BigGAN backbone, in the \myres{64} resolution, we followed the experimental setup of~\cite{noroozi2020self}, where: $BS\!=\!256$, $CH\!=\!64$, $D_{lr}\!=\!G_{lr}\!=\!1\mathrm{e}{-4}$ and found that, although the unconditional BigGAN baseline achieves better metrics with $D_{updates}\!=\!2$, \ours and BigGAN do so with $D_{updates}\!=\!1$. Note that we explored additional configurations such as increasing $BS$ or $CH$ but did not observe any improvement upon the aforementioned setup. In both the \myres{128} and \myres{256} resolutions, BigGAN hyperparameters were borrowed from~\cite{brock2018large}. For \ours, we explored $D_{lr}, G_{lr} \in \{4\mathrm{e}{-4}, 2\mathrm{e}{-4}, 1\mathrm{e}{-4}, 4\mathrm{e}{-5}, 2\mathrm{e}{-5}, 1\mathrm{e}{-5}\}$ and $D_{updates} \in \{1, 2\}$. 
For \myres{128}, we used $BS\!=\!2,\!048$, $CH\!=\!96$ (as in~\cite{brock2018large}), $D_{lr}\!=\!1\mathrm{e}{-4}$, $G_{lr}\!=\!4\mathrm{e}{-5}$ and $D_{updates}\!=\!1$. 
For \myres{256}, we set $BS\!=\!2,\!048$ and $CH\!=\!64$ (half capacity, therefore faster training) for both BigGAN and \ours, and used $D_{lr}\!=\!G_{lr}\!=\!1\mathrm{e}{-4}$ with $D_{updates}\!=\!2$ for \ours. 
When using the StyleGAN2 architecture both as a baseline and as a backbone, we explored $BS \in \{32,64,128,256,512,1,\!024\}$, $D_{lr}, G_{lr} \in \{1\mathrm{e}{-2}, 7\mathrm{e}{-3}, 5\mathrm{e}{-3}, 2.5\mathrm{e}{-3}, 1\mathrm{e}{-4},  5\mathrm{e}{-4}\}$ and $\gamma \in  \{2\mathrm{e}{-1}, 1\mathrm{e}{-2}, 5\mathrm{e}{-2}, 1\mathrm{e}{-1}, 2\mathrm{e}{-1}, 5\mathrm{e}{-1}, 1, 2, 10\}$ and selected $BS\!=\!64$ and $D_{lr}\!=\!G_{lr}\!=\!2.5\mathrm{e}{-3}$ and $\gamma\!=\!5\mathrm{e}{-2}$ for all resolutions.

\paragraph{COCO-Stuff.} When using BigGAN architecture, we explored $BS \in \{128, 256, 512, 2,\!048\}$ and $CH \in \{32, 48, 64\}$  and found $BS\!=\!256$ and $CH\!=\!48$ to be the best choice. We searched for $D_{lr}, G_{lr} \in \{1\mathrm{e}{-3}, 4\mathrm{e}{-4}, 1\mathrm{e}{-4}, 4\mathrm{e}{-5}, 1\mathrm{e}{-5}\}$ and $D_{updates} \in \{1,2\}$. For both unconditional BigGAN and \ours, we chose $D_{lr}\!=\!4\mathrm{e}{-4}$ and $G_{lr}\!=\!1\mathrm{e}{-4}$ in \myres{128} and $D_{lr}\!=\!G_{lr}\!=\!1\mathrm{e}{-4}$ in \myres{256}. 
For both resolutions, unconditional BigGAN uses $D_{updates}\!=\!2$ and \ours, $D_{updates}\!=\!1$. When using StyleGAN2 architecture, we tried several learning rates $D_{lr}, G_{lr} \in \{1\mathrm{e}{-3}, 1.5\mathrm{e}{-3}, 2\mathrm{e}{-3}, 2.5\mathrm{e}{-3}, 3\mathrm{e}{-3}\}$ in combination with $\gamma \in  \{2\mathrm{e}{-1}, 1\mathrm{e}{-2}, 5\mathrm{e}{-2}, 1\mathrm{e}{-1}, 2\mathrm{e}{-1}, 5\mathrm{e}{-1}, 1, 2, 10\}$. For the unconditional StyleGAN2 and IC-GAN trained at resolution \myres{128}, we chose $D_{lr}\!=\!G_{lr}\!=\!2.5\mathrm{e}{-3}$ with $\gamma\!=\!5\mathrm{e}{-2}$. At resolution \myres{256}, we found that $D_{lr}\!=\!G_{lr}\!=\!3\mathrm{e}{-3}$ with $\gamma\!=\!0.5$ were optimal for IC-GAN while we obtained $D_{lr}\!=\!G_{lr}\!=\!2\mathrm{e}{-3}$ with $\gamma\!=\!2\mathrm{e}{-1}$ for the unconditional StyleGAN.  

\paragraph{ImageNet-LT.} We explored $BS \in \{128, 256, 512, 1,\!024, 2,\!048\}$ and $CH \in \{48, 64, 96\}$ and found $BS\!=\!128$ and $CH\!=\!64$ to be the best configuration. We explored $D_{lr}, G_{lr} \in \{1\mathrm{e}{-3}, 4\mathrm{e}{-4}, 1\mathrm{e}{-4}, 4\mathrm{e}{-5}, 1\mathrm{e}{-5}\}$ and $D_{updates} \in \{1, 2\}$. In \myres{64}, we used $D_{lr}\!=\!1\mathrm{e}{-3}$, $G_{lr}\!=\!1\mathrm{e}{-5}$ and $D_{updates}\!=\!1$ for both BigGAN and \ours setup. In \myres{128} and \myres{256}, we used  $D_{lr}\!=\!G_{lr}\!=\!1\mathrm{e}{-4}$ and $D_{updates}\!=\!2$.

\paragraph{Data augmentation.} We use horizontal flips to augment the real data fed to the discriminator in all experiments, unless stated otherwise. For COCO-Stuff and \ImNet-LT, we found that using translations with the DiffAugment framework~\cite{zhao2020differentiable} improves FID scores, as the number of training samples is significantly smaller than \ImNet (5\% and 10\% the size of \ImNet, respectively). However, we did not see any improvement in \ImNet dataset and therefore we do not use DiffAugment in our \ImNet experiments.
For \ImNet and COCO-Stuff, we augment the conditioning instance features $\mathbf{h}_i$ by horizontally flipping all data samples $\mathbf{x}_i$ and obtaining a new $\mathbf{h}_i$ from the flipped image, unless stated otherwise in the tables. This effectively doubles the number of conditionings available at training time, which have the same sample neighborhood $\mathcal{A}_i$ as their non-flipped versions. We tried applying this augmentation technique to \ImNet-LT but found that it degraded the overall metrics, possibly due to the different feature extractor used in these experiments. We hypothesize that the benefits of this technique are dependent on the usage of horizontal flips during the training stage of the feature extractor.
As seen in Table~\ref{table:coco_da_instance}, using data augmentation in the conditioning instance features slightly improves the results for \ours both when coupled with BigGAN and StyleGAN2 backbones in COCO-Stuff.

\paragraph{Compute resources.} We used NVIDIA V100 32GB GPUs to train our models. Given that we used different batch sizes for different experiments, we adapted the resources to each dataset configuration. In particular, \ImNet \myres{64} models were trained using 1 GPU, whereas \ImNet \myres{128} and \myres{256} models required 32 GPUs. \ImNet-LT \myres{64}, \myres{128} and \myres{256} used 1, 2 and 8 GPUs each, respectively. Finally, COCO-Stuff \myres{128} and \myres{256} required 4 and 16 GPUs, respectively, when using the BigGAN backbone, but required 2 and 4 GPUs when leveraging StyleGAN2.

 \begin{table}[h]
\centering
\footnotesize
\caption{Comparison between \ours with and without data augmentation using the  COCO-Stuff dataset. $^\dagger$: 50\% chance of horizontally flipping data samples $\mathbf{x}_i$ to later obtain $\mathbf{h}_i$. The backbone for each \ours is indicated with the number of parameters between parentheses. To compute FID in the training split, we use a subset of $1,\!000$ training instance features (selected with k-means) as conditionings.}
 \begin{tabular}{@{}lccccc@{}}
\toprule
 & \rotatebox[origin=c]{0}{\textbf{Backbone (M)}}  & \multicolumn{4}{c}{$\downarrow$\textbf{FID}} \\
 & &\rotatebox[origin=c]{0}{\textbf{train}}  &  \rotatebox[origin=c]{0}{\textbf{eval}}   &
 \rotatebox[origin=c]{0}{\textbf{eval seen}}
&  \rotatebox[origin=c]{0}{\textbf{eval unseen}}  \\ 
  \midrule
 \multicolumn{3}{l}{128x128} \\ \midrule
\ours & BigGAN (22) & 18.0 $\pm$ 0.1 & 45.5 $\pm$ 0.7 & 85.0 $\pm$ 1.1 & 60.6 $\pm$ 0.9 \\
\ours$^\dagger$ & BigGAN (22) & \textbf{16.8} $\pm$ 0.1 & \textbf{44.9} $\pm$ 0.5 & \textbf{81.5} $\pm$ 1.3 & \textbf{60.5} $\pm$ 0.5 \\
\midrule
\ours  & StyleGAN2 (24) & 8.9 $\pm$ 0.0 & 36.2 $\pm$ 0.2 & 74.3 $\pm$ 0.8 & 50.8 $\pm$ 0.3\\
\ours$^\dagger$ & StyleGAN2 (24) & \textbf{8.7} $\pm$ 0.0  &  \textbf{35.8} $\pm$ 0.1 & \textbf{74.0} $\pm$ 0.7 & \textbf{50.5} $\pm$ 0.6\\
\midrule
\multicolumn{3}{l}{256x256} \\ \midrule
\ours & BigGAN (26) & 25.6 $\pm$ 0.1 & 53.2 $\pm$ 0.3 & 91.1 $\pm$ 3.3 & \textbf{68.3} $\pm$ 0.9 \\
\ours$^\dagger$ & BigGAN (26) & \textbf{24.6} $\pm$ 0.1 & \textbf{53.1} $\pm$ 0.4 & \textbf{88.5} $\pm$ 1.8 & 69.1 $\pm$ 0.6 \\
\midrule
\ours & StyleGAN2 (24.5) & 10.1 $\pm$ 0.0 & 41.8 $\pm$ 0.3 & 78.5 $\pm$ 0.9 & 57.8 $\pm$ 0.6 \\
\ours$^\dagger$ & StyleGAN2 (24.5) & \textbf{9.6} $\pm$ 0.0 & \textbf{41.4} $\pm$ 0.2 & \textbf{76.7} $\pm$ 0.6 & \textbf{57.5} $\pm$ 0.5 \\
\bottomrule
\end{tabular}
 \label{table:coco_da_instance}
 \end{table}

\section{Additional metrics: Precision and Recall}
\label{app:pr_metrics}
As additional measures of visual quality and diversity, we compute Precision (P) and Recall (R)~\cite{kynkaanniemi2019improved} in Table~\ref{table:pr_table}. Results are provided on the \ImNet dataset, following the experimental setup proposed in~\cite{naeem2020reliable}. 
By inspecting the results, we conclude that IC-GAN obtains better Recall (and therefore more diversity) than all the baselines in both the unlabeled and labeled settings, when selecting 10,000 random instances from the training set. Moreover,  when selecting 1,000 instances with k-means, which is the standard experimental setup we used across the paper,  we obtain higher Precision (as a measure of visual quality) than the other baselines in the unlabeled setting. 
In the labeled setting, the Precision is also higher than the one of BigGAN for 64x64 while being  lower for 128$\times$128 and 256$\times$256.

\begin{table}
\caption{Results for \ImNet in terms of Precision (P) and Recall (R)~\cite{kynkaanniemi2019improved} (bounded between 0 and 100), using 10,000 real and generated images. "Instance selection", only used for \ours, indicates whether 1,000 conditioning instances are selected with k-means (k-means 1,000) or 10,000 conditioning instances are sampled uniformly (random 10,000) from the training set to obtain 10,000 generated images in both cases.
 *: Generated images obtained with the paper's opensourced code.
\looseness-1}\label{table:pr_table}
\footnotesize
\centering
\begin{tabular}{@{}lllll@{}}
\toprule
\textbf{Method}
&  \textbf{Res.} & \textbf{Instance selection} & $\uparrow$\textbf{P} & $\uparrow$\textbf{R} \\  \midrule
\textit{Unlabeled setting}\\ 
\midrule
Uncond.\ BigGAN & 64 & - & 69.6 $\pm$ 1.0 & 63.1 $\pm$ 0.0\\
\ours & 64 & k-means 1,000 & \textbf{74.2} $\pm$ 0.8 & 60.2 $\pm$ 0.6 \\
\ours & 64 & random 10,000 & 67.5 $\pm$ 0.4 & \textbf{68.6} $\pm$ 0.5 \\
\midrule
Self-cond. GAN~\cite{liu2020diverse}* & 128 & - & 66.3 $\pm$ 0.5 & 48.4 $\pm$ 0.8 \\ 
\ours & 128 & k-means 1,000 & \textbf{78.2} $\pm$ 0.8 & 55.6 $\pm$ 0.9 \\ 
\ours & 128 & random 10,000 & 71.7 $\pm$ 0.3 & \textbf{69.7} $\pm$ 0.9 \\ 
\midrule
\ours & 256 & k-means 1,000 & \textbf{77.7} $\pm$ 0.5 & 54.3 $\pm$ 0.7 \\ 
\ours & 256 & random 10,000 & 70.4 $\pm$ 0.7 & \textbf{68.9} $\pm$ 0.3 \\ 
\midrule
\textit{Labeled setting}\\ 
\midrule
BigGAN~\cite{brock2018large} & 64 & - & 72.8 $\pm$ 0.4 & 68.6 $\pm$ 0.6\\
Class-conditional \ours & 64 & k-means 1,000 & \textbf{76.6} $\pm$ 0.7 & 67.5 $\pm$ 0.8 \\ 
Class-conditional \ours & 64 & random 10,000 & 69.6 $\pm$ 0.9 & \textbf{74.5} $\pm$ 0.8 \\ 
\midrule
BigGAN~\cite{brock2018large} & 128 & - & \textbf{83.2} $\pm$ 0.7 & 64.2 $\pm$ 0.7\\
Class-conditional \ours & 128 & k-means 1,000 & 78.8 $\pm$ 0.3 & 64.3 $\pm$ 0.7 \\ 
Class-conditional \ours & 128 & random 10,000 & 72.2 $\pm$ 0.4 & \textbf{73.6} $\pm$ 0.5 \\ 
\midrule
BigGAN~\cite{brock2018large} & 256 & - & \textbf{83.9} $\pm$ 0.6 & 70.2 $\pm$ 0.7\\
Class-conditional \ours & 256 & k-means 1,000 & 82.2 $\pm$ 0.3 & 70.4 $\pm$ 0.3 \\ 
Class-conditional \ours & 256 & random 10,000 & 73.9 $\pm$ 0.6 & \textbf{79.3} $\pm$ 0.2 \\ 

\bottomrule
\end{tabular}
\end{table}

\section{Comparison between StyleGAN2 and BigGAN backbones on \ImNet}
\label{app:stylegan_biggan}
We present additional experiments with \ours using the StyleGAN2 backbone in \ImNet in Table~\ref{table:unsup_in_stylegan}, comparing them to StyleGAN2 across all resolutions. \ours with a StyleGAN2 backbone obtains better FID and IS than StyleGAN2 across all resolutions, further supporting that \ours does not depend on a specific backbone, as already shown in the COCO-Stuff dataset in Table~\ref{table:coco_quantitative}. StyleGAN2, despite being designed for unconditional generation, is outperformed by the unconditional counterpart of BigGAN, that uses a single label for the entire dataset, in \ImNet. We suspect that there might be some biases introduced in the architecture at design time, as BigGAN was proposed for \ImNet and StyleGAN2 was tested on datasets with human faces, cars, and dogs, generally with presumably lower complexity and less number of data points than \ImNet. This intuition is further supported by StyleGAN2 improving over the BigGAN backbone in the COCO-Stuff experiments in Table~\ref{table:coco_quantitative}, as this dataset is much smaller than ImageNet and contains a lot of images where people are depicted. Interestingly, we qualitatively found that people and their faces are better generated with a StyleGAN2 backbone rather than the BigGAN one when trained on COCO-Stuff.

\begin{table}
\caption{Results for  \ImNet in unlabeled setting, comparing BigGAN and StyleGAN backbones.
For fair comparison with~\cite{noroozi2020self} at $64\!\times\!64$ resolution, we trained an unconditional BigGAN model and report the non-official FID and IS scores -- computed with Pytorch rather than  TensorFlow -- indicated with *. $^\dagger$: increased parameters to match \ours capacity. DA: 50\% horizontal flips in real and fake samples (\textbf{d}), and conditioning instances (\textbf{i}). $ch \times$: Channel multiplier that affects network width.
\looseness-1}\label{table:unsup_in_stylegan}
\footnotesize
\centering
\begin{tabular}{@{}llll@{}}
\toprule
\textbf{Method}
&  \textbf{Res.} & $\downarrow$\textbf{FID} & $\uparrow$\textbf{IS} \\  \midrule
Uncond.\ BigGAN$^\dagger$ & 64 & 16.9* $\pm$ 0.0 & 14.6* $\pm$ 0.1 \\
\textbf{StyleGAN2} + DA (\textbf{d}) & 64 & 12.4* $\pm$ 0.0 & 15.4* $\pm$ 0.0 \\ 
\textbf{\ours (BigGAN)} + DA (\textbf{d},\textbf{i}) & 64 & 9.2* $\pm$ 0.0 & \textbf{23.5}* $\pm$ 0.1 \\ 
\textbf{\ours (StyleGAN2)} + DA (\textbf{d},\textbf{i}) & 64 &  \textbf{8.5}* $\pm$ 0.0 & \textbf{23.5}* $\pm$ 0.1 \\ 
\midrule
Uncond.\ BigGAN~\cite{pmlr-v97-lucic19a} & 128 & 25.3 & 20.4 \\
\textbf{StyleGAN2} + DA (\textbf{d}) & 128 &  27.8 $\pm$ 0.1 & 18.8 $\pm$ 0.1 \\ 
\textbf{\ours (BigGAN)} + DA (\textbf{d},\textbf{i}) & 128 & \textbf{11.7} $\pm$ 0.0 & \textbf{48.7} $\pm$ 0.1 \\ 
\textbf{\ours (StyleGAN2)} + DA (\textbf{d},\textbf{i}) & 128 & 15.2 $\pm$ 0.1 & 38.3 $\pm$ 0.2 \\ 
\midrule 

\textbf{StyleGAN2} + DA (\textbf{d}) & 256 &  41.3 $\pm$ 0.1 & 19.7 $\pm$ 0.1 \\ 
\textbf{\ours (BigGAN)} ($ch\times64$) + DA (\textbf{d},\textbf{i}) & 256 &  17.4 $\pm$ 0.1 & 53.5 $\pm$ 0.5 \\
\textbf{\ours (BigGAN)} ($ch\times96$) + DA (\textbf{d}) & 256 &  \textbf{15.6} $\pm$ 0.1 & \textbf{59.0} $\pm$ 0.4 \\
\textbf{\ours (StyleGAN2)} + DA (\textbf{d},\textbf{i}) & 256 &  23.1 $\pm$ 0.1 & 42.2 $\pm$ 0.2 \\ 
\bottomrule
\end{tabular}
\end{table}

\section{Additional qualitative results for \ours}
\label{app:add_qualitative}
\begin{figure}
\centering
\begin{subfigure}{1\textwidth}
 \centering
\includegraphics[width=\textwidth]{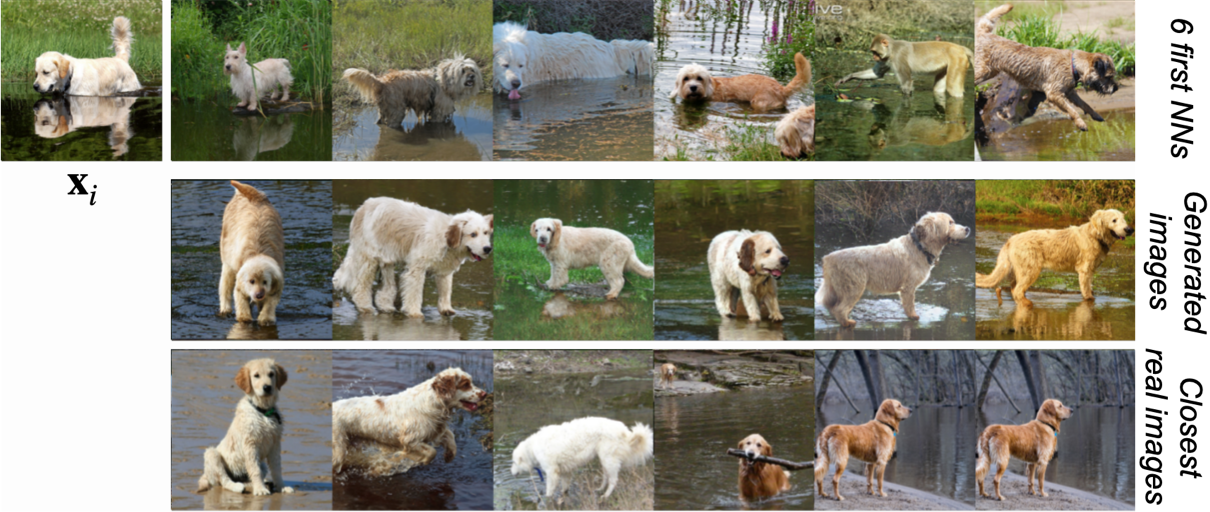}
\end{subfigure}
\begin{subfigure}{1\textwidth}
 \centering
\includegraphics[width=\textwidth]{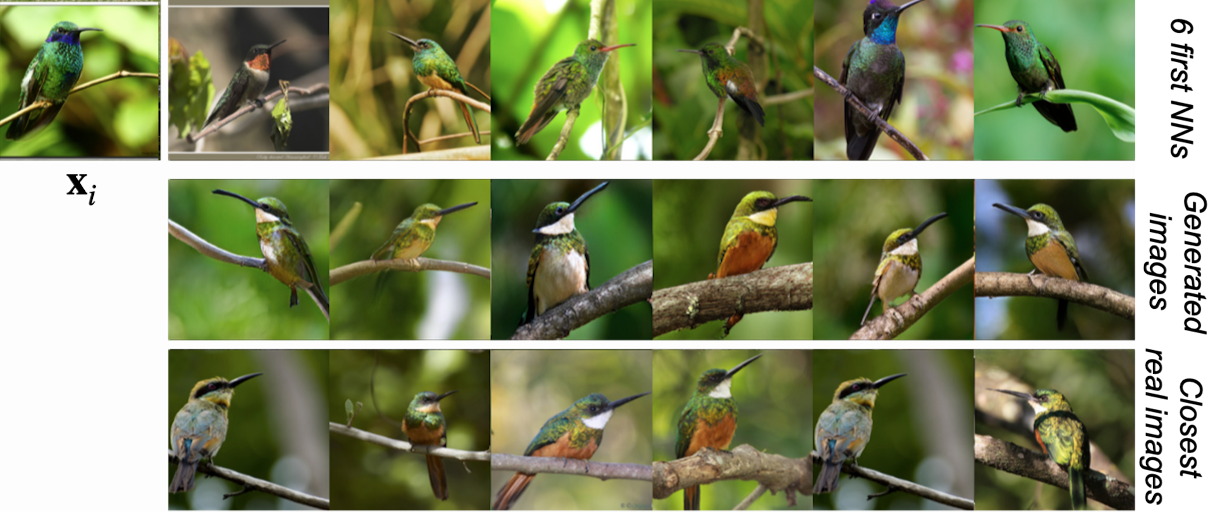}
\end{subfigure}
\begin{subfigure}{1\textwidth}
 \centering
\includegraphics[width=\textwidth]{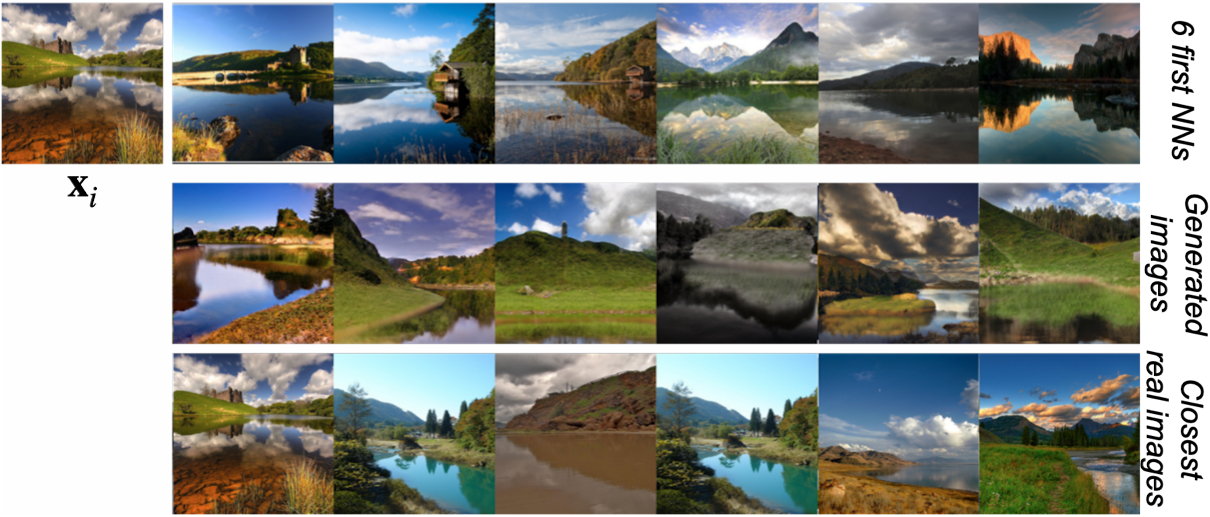}
\end{subfigure}
\caption{Qualitative results on unlabeled \ImNet (\myres{256}). Next to each input sample $\mathbf{x}_i$, used to obtain the instance features $\mathbf{h}_i\,=\,f_\theta(\mathbf{x}_i)$, the six nearest neighbors in the feature space of $f_\theta$ are displayed. Below the neighbors, generated images sampled from \ours with a BigGAN backbone and conditioned on $\mathbf{h}_i$ are depicted. Immediately below the generated images, the closest image in the \ImNet training set is shown for each example (cosine distance in the feature space of $f_\theta$).
}
\label{fig:icgan_qualitative_app}
\end{figure}

\paragraph{Unlabeled \ImNet.} \ours generates high quality and diverse images that generally preserve the semantics and style of the conditioning. Figure~\ref{fig:icgan_qualitative_app} shows three instances -- a golden retriever in the water, a humming bird on a branch, and a landscape with a castle --, followed by their six closest nearest neighbors in the feature space of SwAV~\cite{caron2020unsupervised}, a ResNet50 model trained with self-supervision. 
Note that, although all neighbors contain somewhat similar semantics to those of the instance, the class labels do not always match. For example, one of the nearest neighbors of a golden retriever depicts a monkey in the water. The generated images depicted in Figure~\ref{fig:icgan_qualitative_app} are obtained by conditioning \ours with a BigGAN backbone on the features of the aforementioned instances. These highlight that generated images preserve the semantic content of the conditioning instance (a dog in the water, a bird with a long beak on a branch, and a landscape containing a water body) and present similarities with the real samples in the neighborhood of the instance. In cases such as the conditioning instance featuring a castle, the corresponding generated samples do not contain buildings; this could be explained by the fact that most of its neighbors do not contain castles either. Moreover, the generated images are not mere memorizations of training examples, as shown by the row of images immediately below, nor are they copies of the conditioning instance. 

\begin{figure}
\centering
\begin{subfigure}{0.48\textwidth}
 \centering
\includegraphics[width=\textwidth]{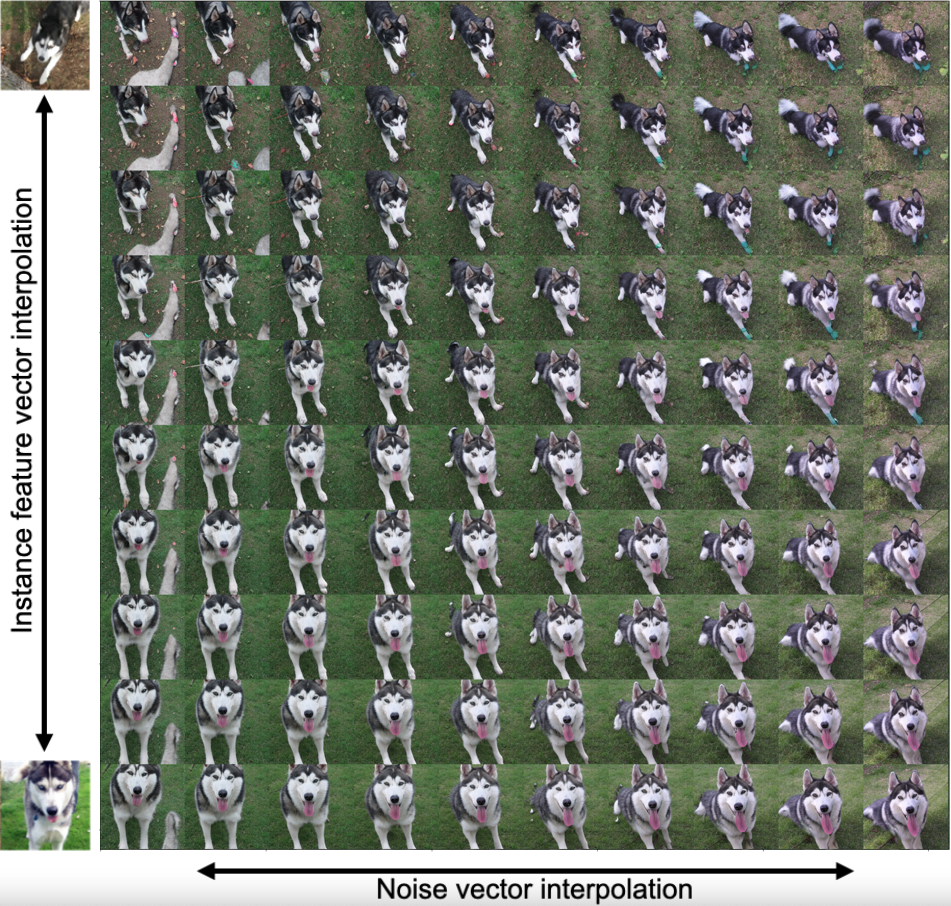}
\end{subfigure}
\begin{subfigure}{0.48\textwidth}
 \centering
\includegraphics[width=\textwidth]{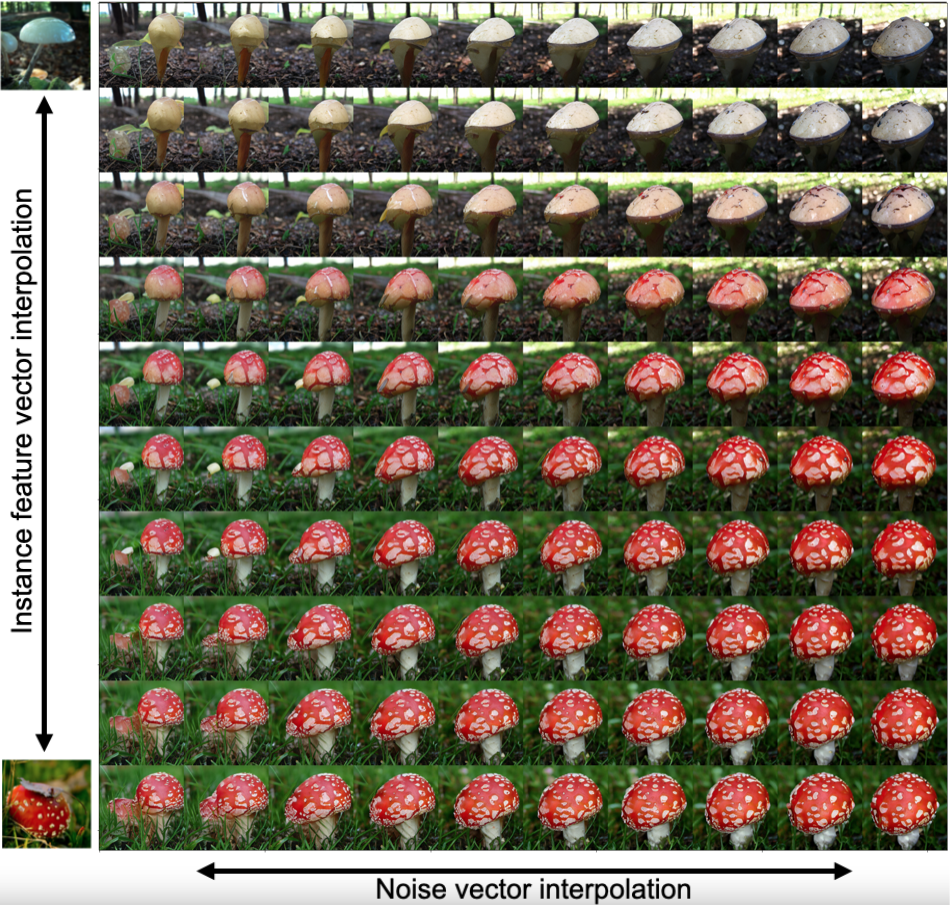}
\end{subfigure}
\begin{subfigure}{0.48\textwidth}
 \centering
\includegraphics[width=\textwidth]{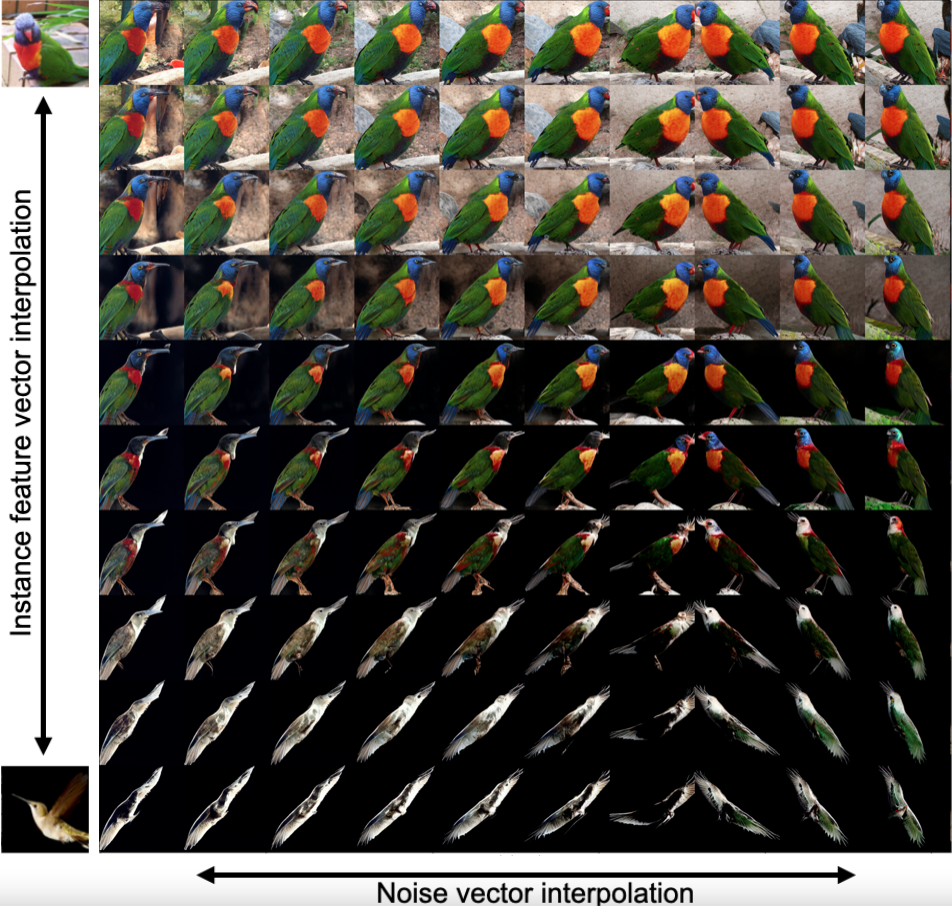}
\end{subfigure}
\begin{subfigure}{0.48\textwidth}
 \centering
\includegraphics[width=\textwidth]{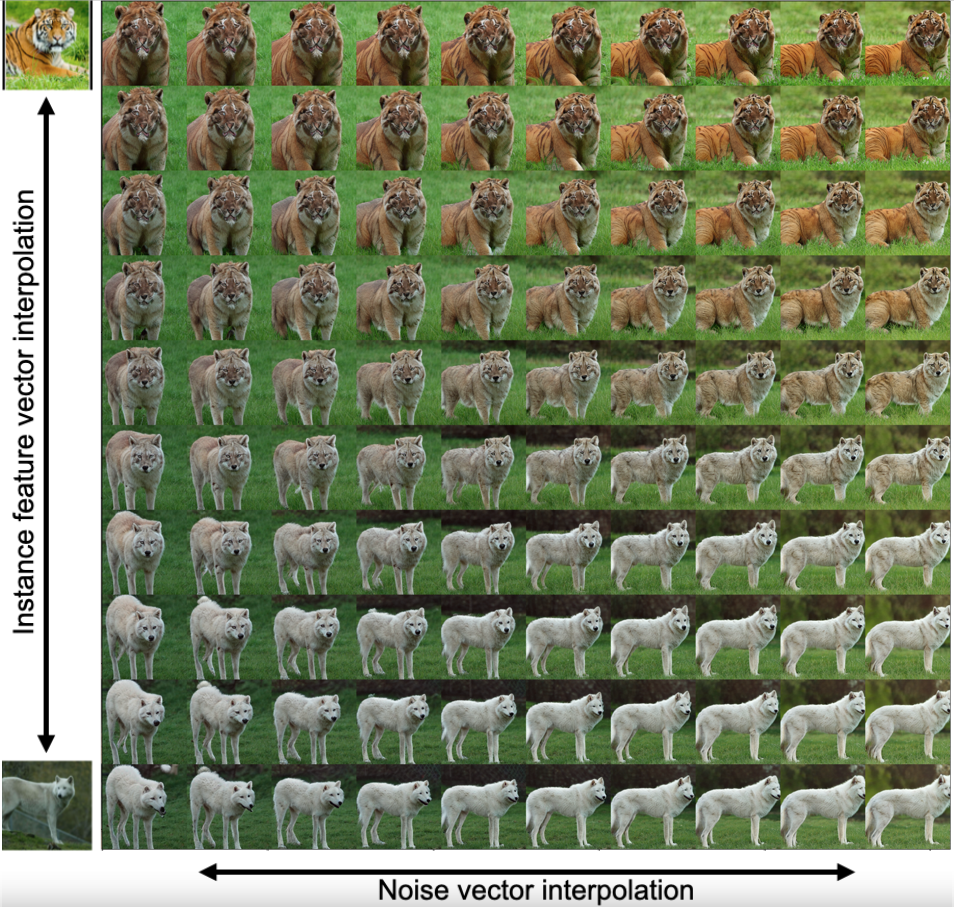}
\end{subfigure}
\caption{Qualitative results on unlabeled \ImNet (\myres{256}) using \ours (BigGAN backbone) and interpolating between two instance feature vector conditionings (vertical axis) and two input noise vectors (horizontal axis).The two images depicted to the left of the generated image grids are used to extract the instance feature vectors used for the interpolation.}
\label{fig:app_interpolation}
\end{figure}

\paragraph{Instance feature vector and noise interpolation.} In Figure~\ref{fig:app_interpolation}, we provide the resulting generated images when interpolating between the instance features of two data samples (vertical axis), shown on the left of each generated image grid, and additionally interpolating between two noise vectors in the horizontal axis. The top left quadrant shows generated images when interpolating between conditioning instance features from the class \textit{husky}. The generated dog changes its fur color and camera proximity according to the instance conditioning. At the top right corner, when interpolating between two \textit{mushroom} instance features, generated images change their color and patterns accordingly. Moreover, in the bottom left quadrant, \textit{lorikeet} instance features are interpolated with flying \textit{hummingbird} instance features, and the generated images change their color and appearance accordingly. Finally, in the bottom right grid, we interpolate instance features from a \textit{tiger} and instance features from a \textit{white wolf}, resulting in different blends between the striped pelt of the tiger and the white fur of the wolf. 

\begin{figure}[h]
\centering
\begin{subfigure}[t]{0.1465\textwidth}
 \centering
\includegraphics[width=\textwidth]{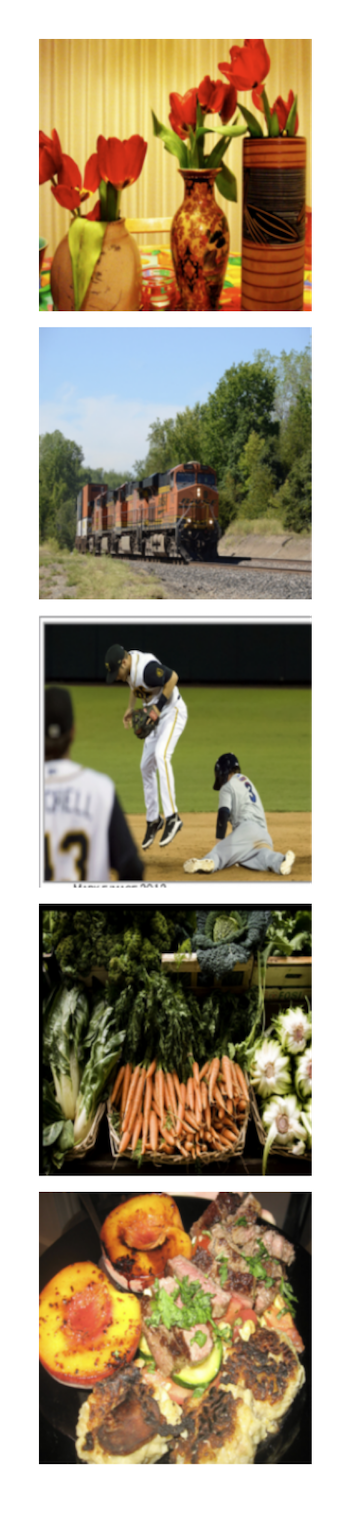}
\caption{$\mathbf{x}_i$ }
\label{subfig:coco_app_input}
\end{subfigure}
\begin{subfigure}[t]{0.26\textwidth}
 \centering
\includegraphics[width=\textwidth]{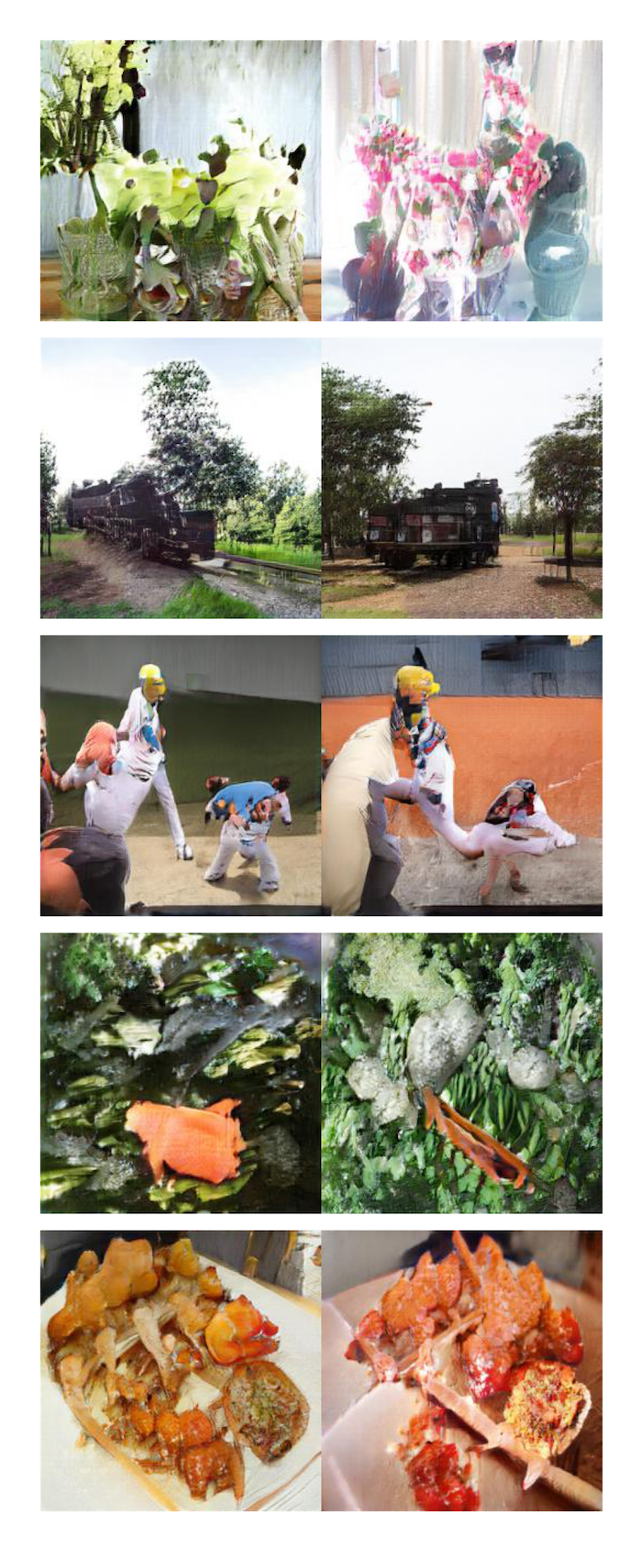}
\caption{LostGANv2~\cite{sun2020learning}}
\label{subfig:coco_app_lg2}
\end{subfigure}
\begin{subfigure}[t]{0.26\textwidth}
 \centering
\includegraphics[width=\textwidth]{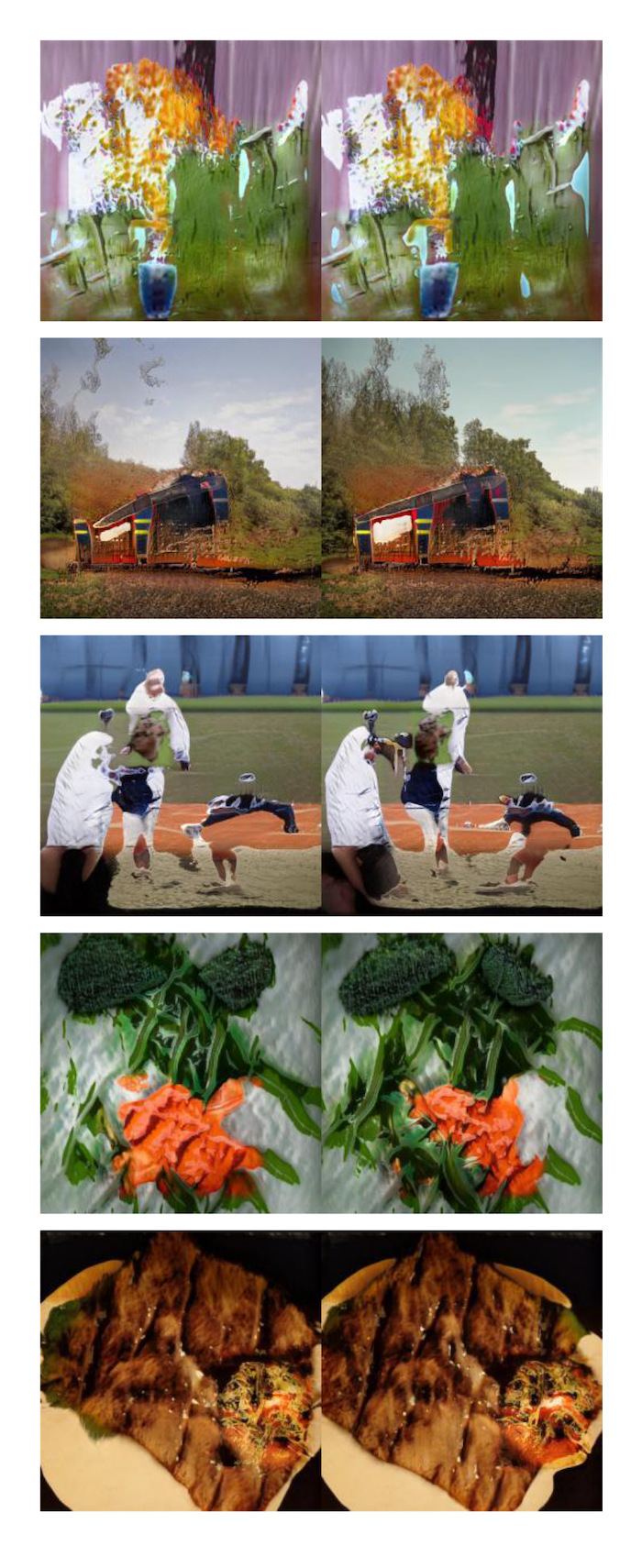}
\caption{OC-GAN~\cite{sylvain2020object} }
\label{subfig:coco_app_ocgan}
\end{subfigure}
\begin{subfigure}[t]{0.26\textwidth}
 \centering
\includegraphics[width=\textwidth]{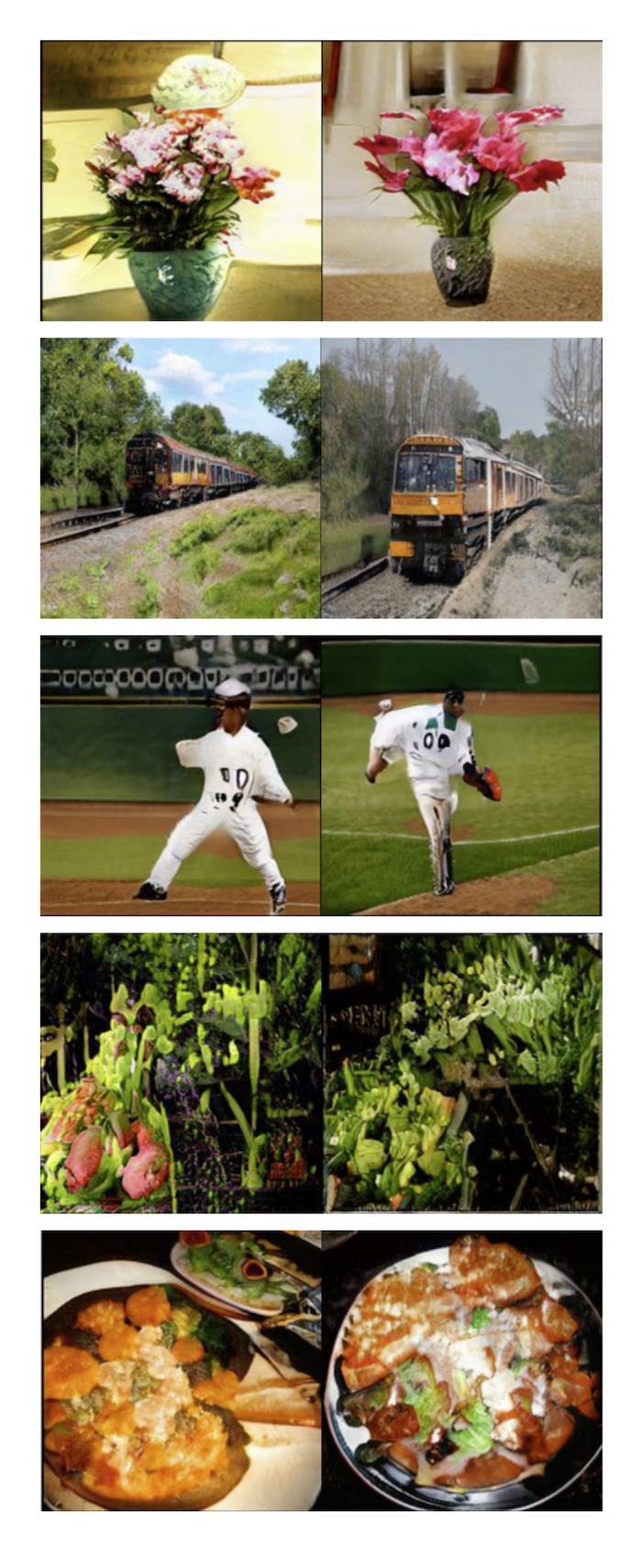}
\caption{\ours (StyleGAN2) }
\label{subfig:coco_app_input2}
\end{subfigure}

\caption{Qualitative comparison for scene generation on \myres{256} COCO-Stuff with other state-of-the-art scene generation methods. (a) Data samples $\mathbf{x}_i$ from which instance features $\mathbf{h}_i\,=\,f(\mathbf{x}_i)$ are obtained for \ours, and labeled bounding box conditionings are obtained for LostGANv2 and OC-GAN. Images generated with two random seeds with (b) LostGANv2~\cite{sun2020learning}, (c) OC-GAN~\cite{sylvain2020object}, (d) \ours (StyleGAN2).
}
\label{fig:coco_qualitative_sota}
\end{figure}

\paragraph{Unlabeled COCO-Stuff.} Training \ours with a StyleGAN2 backbone on COCO-Stuff has resulted in quantitative results that surpass those achieved by the state-of-the-art LostGANv2~\cite{sun2020learning} and OC-GAN~\cite{sylvain2020object}, controllable and conditional complex scene generation pipelines that rely on heavily labeled data (bounding boxes and class labels), tailored intermediate steps and somewhat complex architectures. In Figure~\ref{fig:coco_qualitative_sota}, we compare generated images obtained with LostGANv2 and OC-GAN with those generated by \ours with a StyleGAN2 backbone. Note that the two former methods use a bounding box layout with class labels as a conditioning, while we condition on the features extracted from the real samples $\mathbf{x}_i$ depicted in Figure~\ref{subfig:coco_app_input}. We compare the generations obtained with two random seeds for all methods, and observe that \ours generates higher quality images in all cases, especially for the top three instances. Moreover, the diversity in the generations using two random seeds for LostGANv2 and OC-GAN is lower than for \ours. This is not surprising, as the former methods are restricted by their bounding box layout conditioning that specifies the number of objects, their classes and their expected positions in the generated images. By contrast, \ours conditions on an instance feature vector, which does not require any object label, cardinality or position to be satisfied, allowing more freedom in the generations.

\begin{figure}
\centering
\begin{subfigure}{1\textwidth}
 \centering
\includegraphics[width=\textwidth]{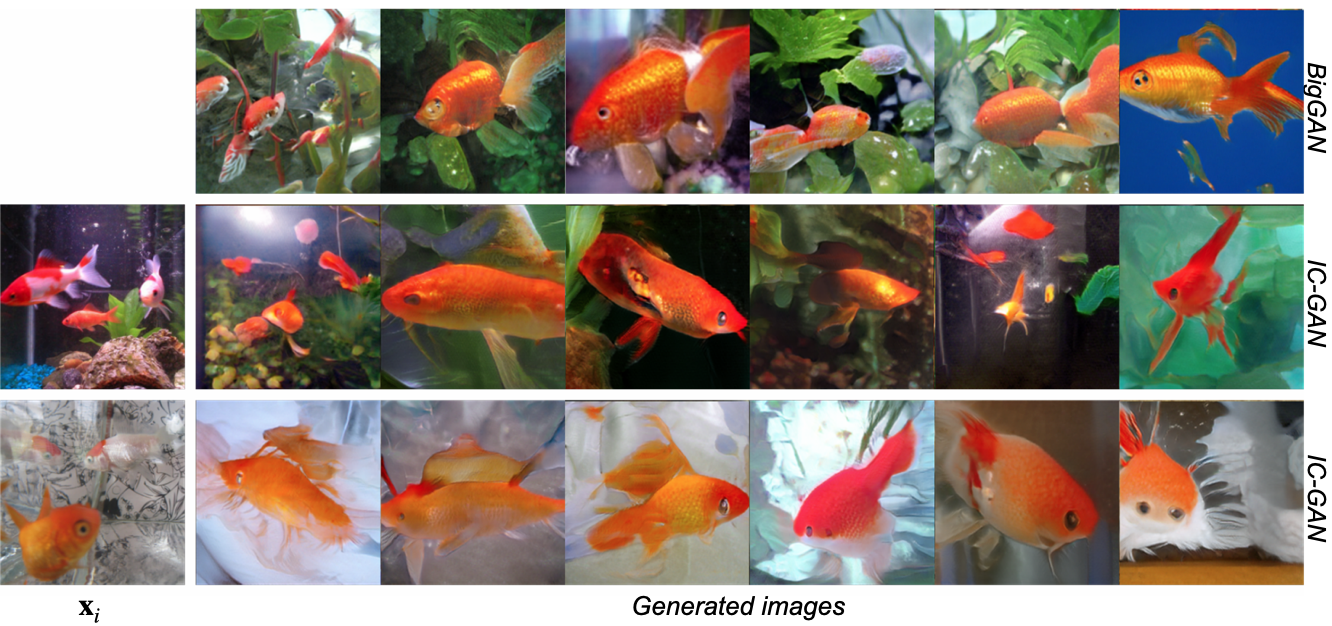}
\caption{Class label \textit{Goldfish}}
\label{app:goldfish}
\end{subfigure}
\begin{subfigure}{1\textwidth}
 \centering
\includegraphics[width=\textwidth]{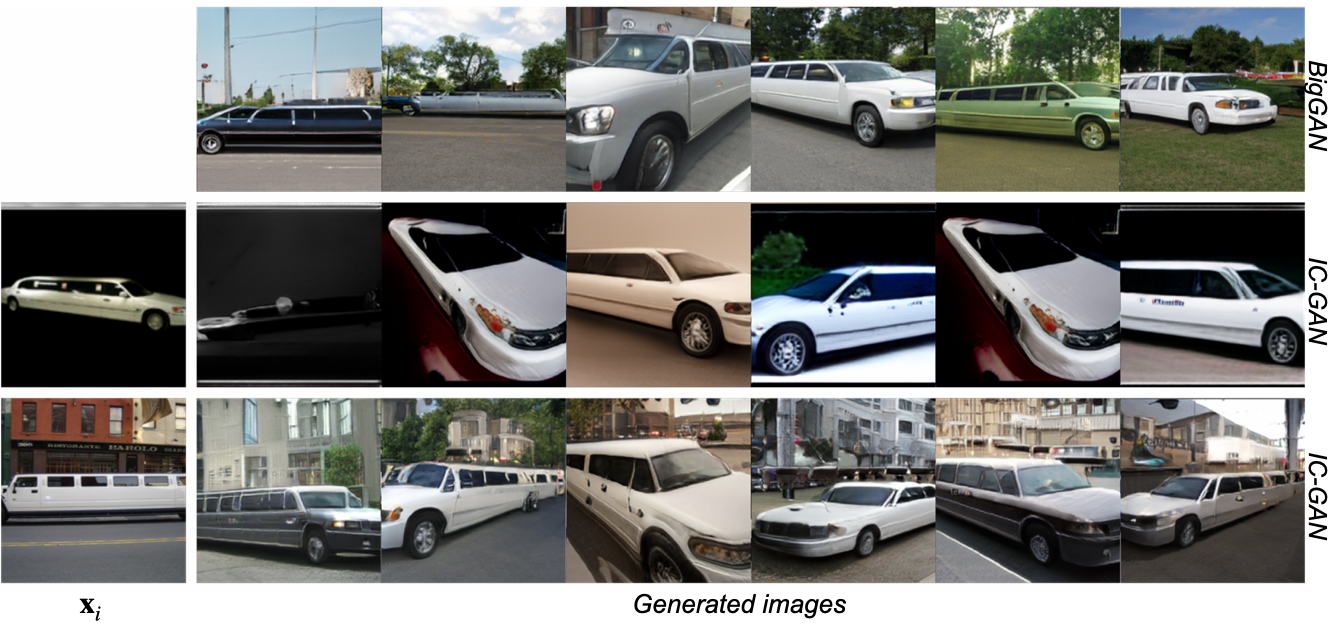}
\caption{Class label \textit{Limousine}}
\label{app:limousine}
\end{subfigure}
\begin{subfigure}{1\textwidth}
 \centering
\includegraphics[width=\textwidth]{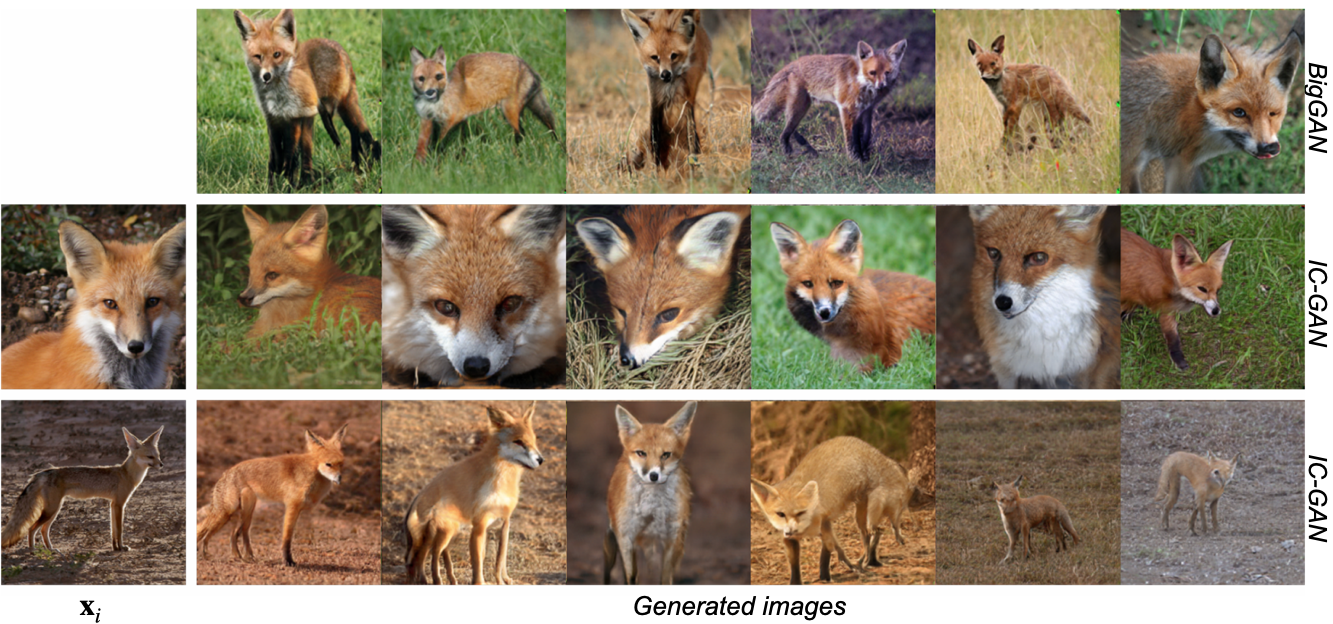}
\caption{Class label \textit{Red fox}}
\label{app:red_fox}
\end{subfigure}
\caption{Qualitative results in \myres{256} \ImNet. For each class, generated images with BigGAN are presented in the first row, while the second and third row show generated images using class-conditional \ours with a BigGAN backbone, conditioned on the instance feature extracted from the data sample to their left ($\mathbf{x}_i$) and their corresponding class.}
\label{fig:icgan_cc_qualitative_app}
\end{figure}

\paragraph{\ImNet.} Class-conditional \ours with a BigGAN backbone has shown comparable quantitative results to those of BigGAN for \myres{256} resolution in Subsection~\ref{sec:supervised}. In Figure~\ref{fig:icgan_cc_qualitative_app}, we can qualitatively compare BigGAN ($ch\times64$) (first rows) and \ours ($ch\times64$) (second and third rows), for three class labels: \textit{goldfish}, \textit{limousine} and \textit{red fox}. By visually inspecting the generated images, we can observe that the generation quality is similar for both BigGAN and \ours in these specific cases. Moreover, \ours allows controllability of the semantics by changing the conditioning instance features. For instance, changing the background in which the goldfish are swimming into lighter colors in Figure~\ref{app:goldfish}, generating limousines in generally dark and uniform backgrounds or, instead, in an urban environment with a road and buildings (Figure~\ref{app:limousine}), or generating red foxes with a close up view or with a full body shot as seen in Figure~\ref{app:red_fox}.

\begin{figure}
\centering
\begin{subfigure}{0.90\textwidth}
 \centering
\includegraphics[width=\textwidth]{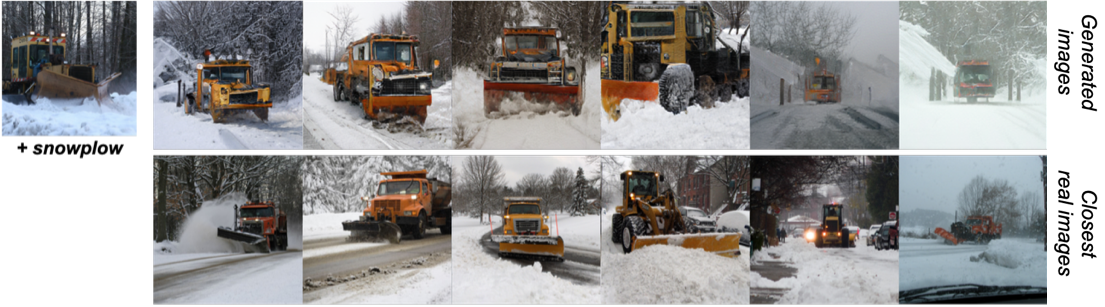}
\end{subfigure}
\begin{subfigure}{0.90\textwidth}
 \centering
\includegraphics[width=\textwidth]{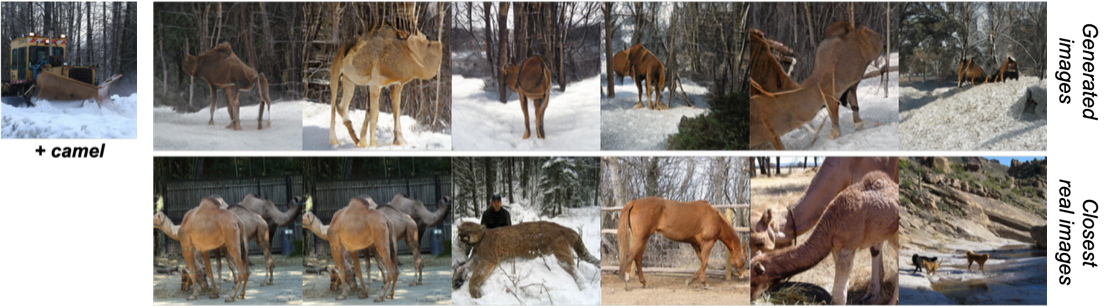}
\end{subfigure}
\begin{subfigure}{0.90\textwidth}
 \centering
\includegraphics[width=\textwidth]{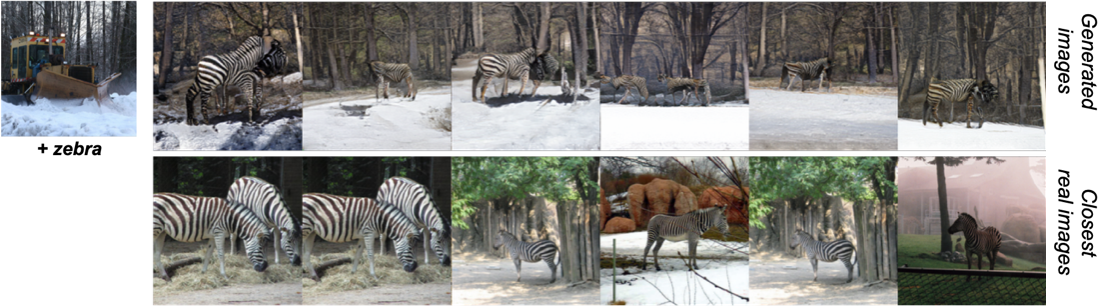}
\end{subfigure}
\begin{subfigure}{0.90\textwidth}
 \centering
\includegraphics[width=\textwidth]{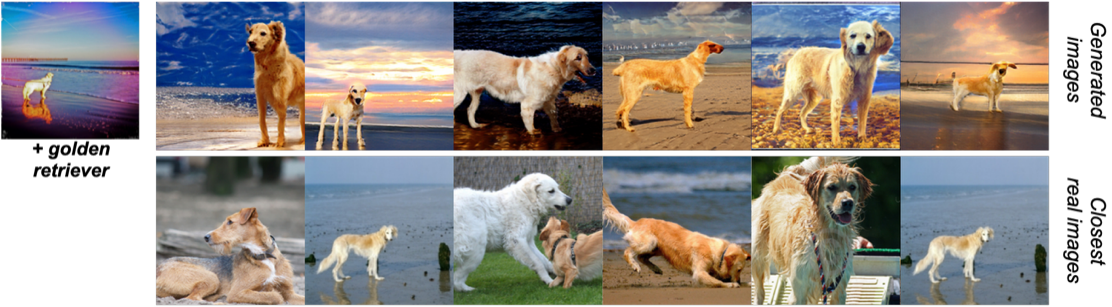}
\end{subfigure}
\begin{subfigure}{0.90\textwidth}
 \centering
\includegraphics[width=\textwidth]{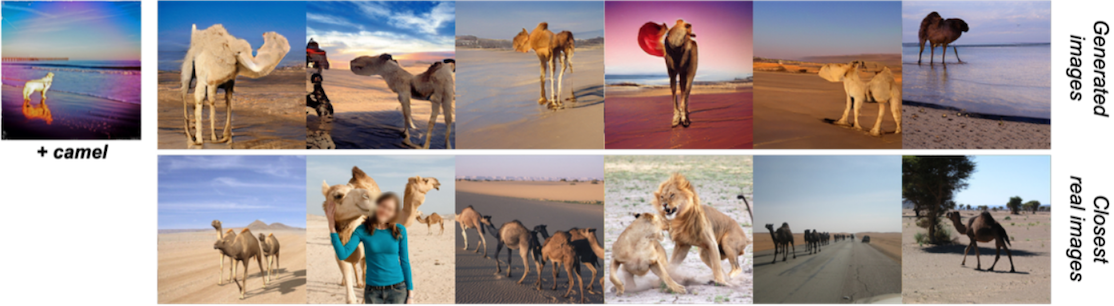}
\end{subfigure}
\begin{subfigure}{0.90\textwidth}
 \centering
\includegraphics[width=\textwidth]{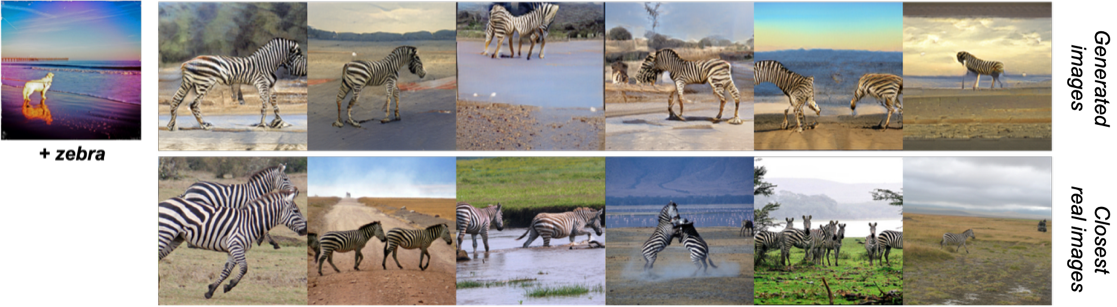}
\end{subfigure}

\caption{Generated \myres{256} images with a class-conditional \ours (BigGAN backbone) trained on \ImNet. Next to each data sample $\mathbf{x}_i$, used to obtain the instance features $\mathbf{h}_i\,=\,f_\theta(\mathbf{x}_i)$, we find generated images sampled from \ours using $\mathbf{h}_i$ and six sampled noise vectors. Below the generated images, the closest image in the \ImNet training set are shown (Cosine similarity in the feature space of $f_\theta$).}
\label{fig:icgan_cc_qualitative_swap_app}
\end{figure}

\paragraph{Swapping classes for class-conditional \ours on \ImNet.} 
In Figure~\ref{fig:icgan_cc_qualitative_app}, we show that we can change the appearance of the generated images by leveraging different instances of the same class.
In Figure~\ref{fig:icgan_cc_qualitative_swap_app}, we take a further step and condition on instance features from other classes. More specifically, in Figure~\ref{fig:icgan_cc_qualitative_swap_app} (top), we condition on the instance features of a snowplow in the woods surrounded by snow, and ask to generate snowplows, camels and zebras. Perhaps surprisingly, the generated images effectively get rid of the snowplow, and replace it by camel-looking and zebra-looking objects, respectively, while maintaining a snowy background in the woods. 
Moreover, when comparing the generated images with the closest samples in \ImNet, we see that for generated camels in the snow, the closest images are either a camel standing in dirt or other animals in the snow; for the generated zebras in the snow, we find one sample of a zebra standing in the snow, while others are standing in other locations/backgrounds. 
In Figure~\ref{fig:icgan_cc_qualitative_swap_app} (bottom), we condition on the features of an instance that depicts a golden retriever on a beach with overall purple tones, and ask to generate golden retrievers, camels or zebras. In most cases, generated images contain camels and zebras standing on water, while other generations contain purple or blue tones, similar to the instance used as conditioning. Note that, except one generated zebra image, the closest samples in \ImNet do not depict camels or zebras standing in the water nor on the beach.

\section{Additional off-the-shelf transfer results for \ours}
\label{app:add_transfer}
\paragraph{Is \ours able to shift the generated data distribution by conditioning on different instances?} As discussed in Section~\ref{sec:transfer}, we can transfer an \ours trained on unlabeled \ImNet to COCO-Stuff and obtain better metrics and qualitative results than with the same \ours trained on COCO-Stuff.
We hypothesize that the success of this experiment comes from the flexibility of our conditioning strategy, where the generative model exploits the generalization capabilities of the feature extractor when dealing with unseen instances to shift the distribution of generated images from \ImNet to COCO-Stuff. To test this hypothesis we design the following experiment: we compute FID scores of generated images obtained by conditioning \ours with instance features from either \ImNet or COCO-Stuff and use either COCO-Stuff or \ImNet as a reference dataset to compute FID. In Table~\ref{table:ablation_transfer_shift} (first row) we show that when using COCO-Stuff for both the instance features and the reference dataset, \ours scores 8.5 FID; this is a lower FID score than the 43.6 FID obtained in Table~\ref{table:ablation_transfer_shift} (second row) when conditioning \ours on \ImNet instance features and using COCO-Stuff as reference dataset. Moreover, when using COCO-Stuff instance features and \ImNet as reference dataset, in Table~\ref{table:ablation_transfer_shift} (third row), we obtain 37.2 FID. This shows that, by changing the conditioning instance features, \ours successfully exploits the generalization capabilities of the feature extractor to shift the distribution of generated images to be closer to the COCO-Stuff distribution. Additionally, note that the distance between ImageNet and COCO-Stuff datasets can be quantified with an FID score of 37.2~\footnote{We subsampled $76,\!000$ ground-truth images from ImageNet training set and used all COCO-Stuff training ground-truth images.}.

\begin{table}[h!]
\centering
\footnotesize
\caption{FID scores on COCO-Stuff \myres{128}, when using an \ours trained on ImageNet and tested with instance features from either COCO-Stuff or \ImNet and using either of those datasets as reference. The metrics obtained by sampling $1,\!000$ instance features (k-means) from either ImageNet or COCO, and generating $76,\!000$ samples. As a reference, $76,\!000$ real samples from COCO-Stuff or ImageNet training set are used.}
 \begin{tabular}{@{}lcccc@{}}
\toprule
 & train instance dataset & eval instance dataset & FID reference dataset & $\downarrow$\textbf{FID} \\  \midrule
\ours & ImageNet & COCO-Stuff & COCO-Stuff & \textbf{8.5} $\pm$ 0.1 \\
\ours & ImageNet & ImageNet & COCO-Stuff  & 43.6 $\pm$ 0.1 \\
\ours & ImageNet & COCO-Stuff & ImageNet & 37.2 $\pm$ 0.1 \\ 
\bottomrule
\end{tabular}
 \label{table:ablation_transfer_shift}
 \end{table}
\paragraph{What is being transferred when \ours is conditioned on instances other than the ones in the training dataset?} From the point of view of KDE, what is being transferred is the kernel shape, not the kernel location (that is controlled by instances). The kernel shape is predicted using a generative model from each input instance and we probe the kernel via sampling from the generator. Thus, we transfer a function that predicts kernel shape from a conditioning, and this function seems to be robust to diverse instances as shown in the paper (e.g. see Figure~\ref{subfig:teaser_unsup_transfer} and~\ref{subfig:teaser_class_transfer}). 
Moreover, by visually inspecting the generated images in our transfer experiments, we observed that when transferring an IC-GAN trained on ImageNet to COCO-Stuff, if the model is conditioned on images that contain unseen classes in ImageNet, such as “giraffe”, the model will still generate an animal that would look like a giraffe without the skin patterns and characteristic antennae, because ImageNet contains other animals to draw inspiration from. This suggests that the model generates plausible images that have some similar features to those present in the instance conditioning, but adapting it to the training dataset style. Along these lines, we also observed that in some cases, shapes and other object characteristics from one dataset are transferred to another (ImageNet to COCO-Stuff). Moreover, when we conditioned on instances from Cityscapes, the generated images were rather colorful, resembling more the color palette of ImageNet images rather than the Cityscapes one. 

\begin{figure}
\centering
\begin{subfigure}{1\textwidth}
 \centering
\includegraphics[width=\textwidth]{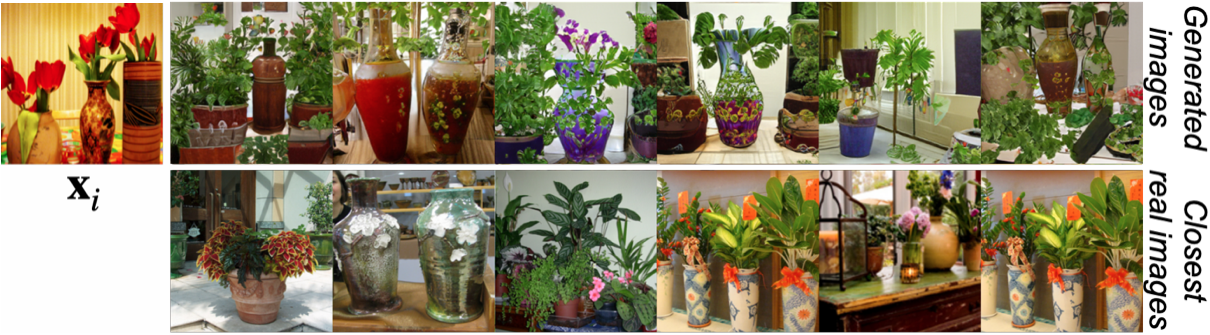}
\end{subfigure}
\begin{subfigure}{1\textwidth}
 \centering
\includegraphics[width=\textwidth]{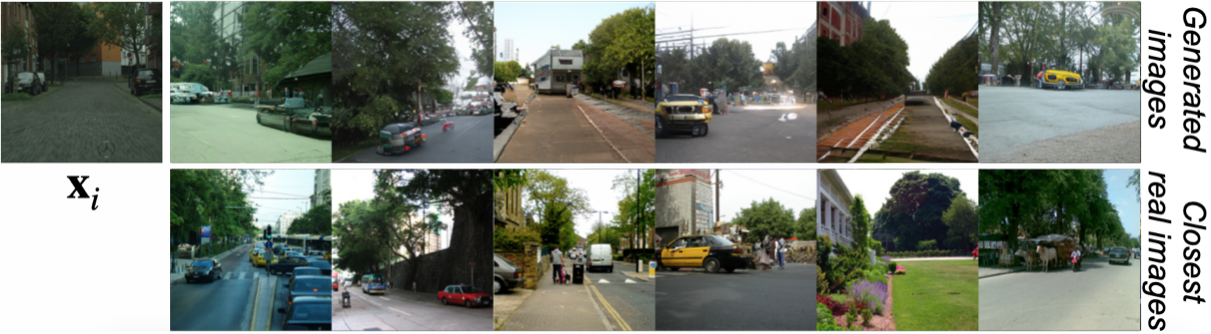}
\end{subfigure}
\begin{subfigure}{1\textwidth}
 \centering
\includegraphics[width=\textwidth]{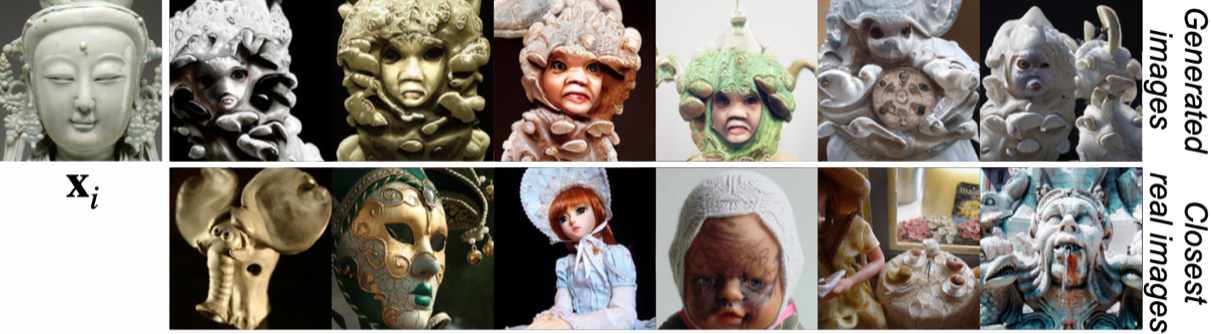}
\end{subfigure}
\begin{subfigure}{1\textwidth}
 \centering
\includegraphics[width=\textwidth]{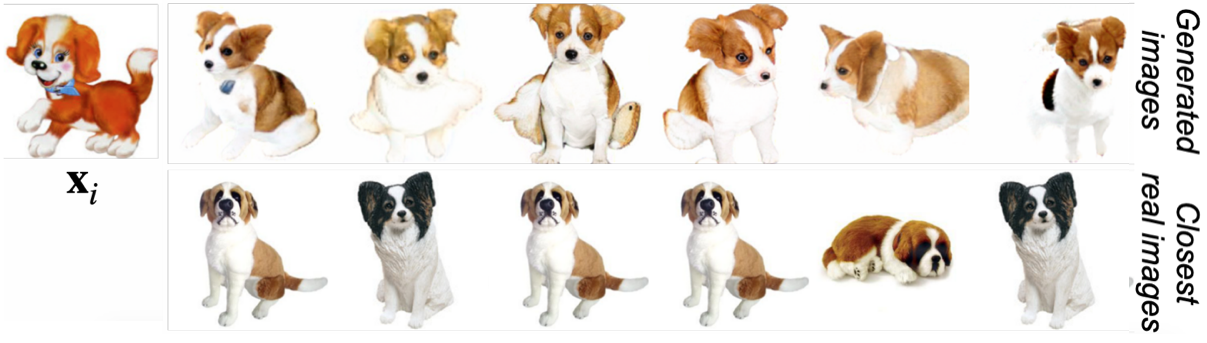}
\end{subfigure}
\begin{subfigure}{1\textwidth}
 \centering
\includegraphics[width=\textwidth]{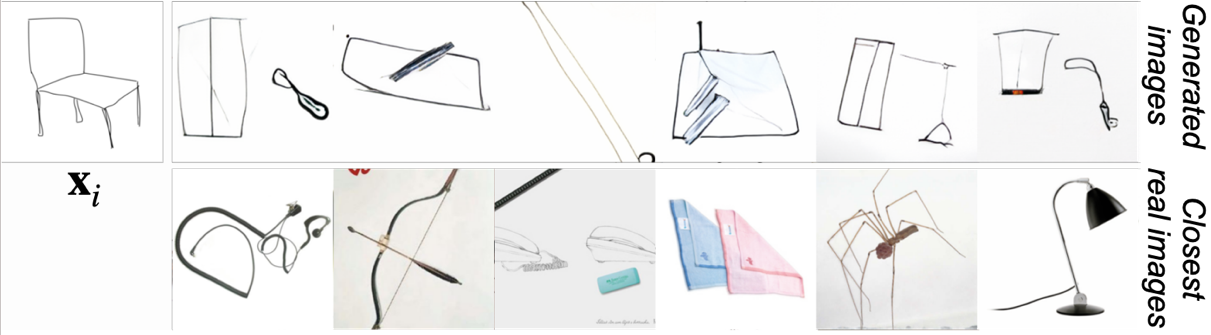}
\end{subfigure}
\caption{Qualitative off-the-shelf transfer results in \myres{256}, using an \ours trained on unlabeled \ImNet and conditioning on unseen instances from other datasets. The instances come from the following datasets (top to bottom): COCO-Stuff, Cityscapes, MetFaces, PACS (cartoons), Sketches. 
Next to each data sample $\mathbf{x}_i$, used to obtain the instance features $\mathbf{h}_i\,=\,f_\theta(\mathbf{x}_i)$, generated images conditioning on $\mathbf{h}_i$ are displayed. Immediately below each generated image, the closest image in the \ImNet training set is displayed (Euclidean distance in the feature space of $f_\theta$).
}
\label{fig:icgan_transfer_app}
\end{figure}

\paragraph{Off-the-shelf transfer results for \ours.} In Figure~\ref{fig:icgan_transfer_app}, we provide additional generated samples and their closest images in the \ImNet training set, when conditioning on unseen instance features from other datasets. 
Generated images often differ substantially from the closest image in \ImNet. Although generated images using a COCO-Stuff and Cityscapes instances may have somewhat similar looking images in \ImNet (for the first and second instances in Figure~\ref{fig:icgan_transfer_app}), the differences accentuate when conditioning on instance features from MetFaces, PACS or Sketch datasets, where, for instance, \ours with a BigGAN backbone generates images resembling sketch strokes in the last row, even if the closest \ImNet samples depict objects that are not sketches.

\begin{figure}
\centering
\begin{subfigure}{1\textwidth}
 \centering
\includegraphics[width=\textwidth]{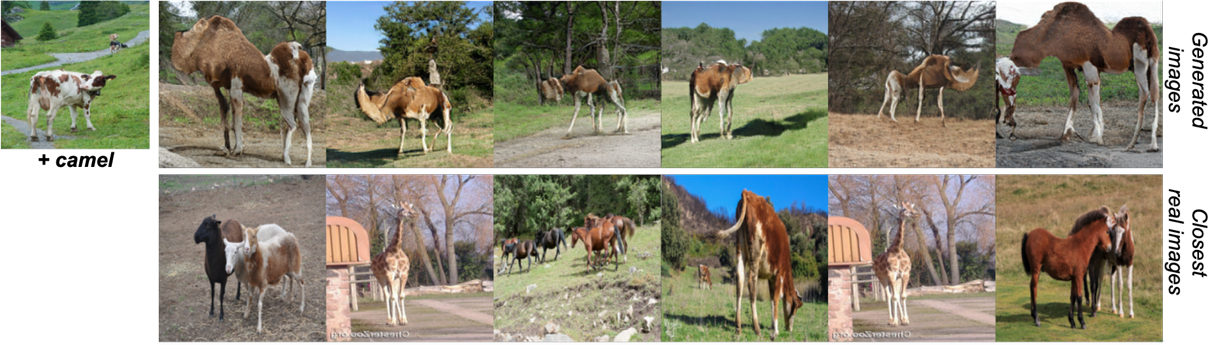}
\end{subfigure}
\begin{subfigure}{1\textwidth}
 \centering
\includegraphics[width=\textwidth]{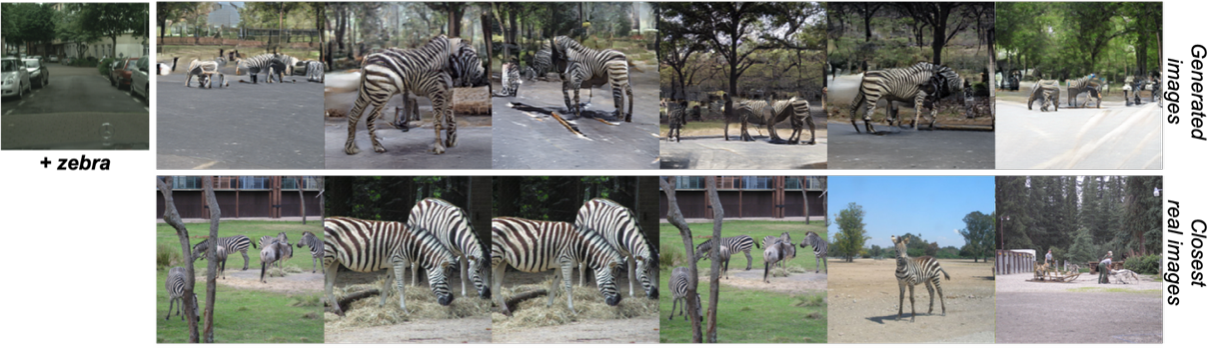}
\end{subfigure}
\begin{subfigure}{1\textwidth}
 \centering
\includegraphics[width=\textwidth]{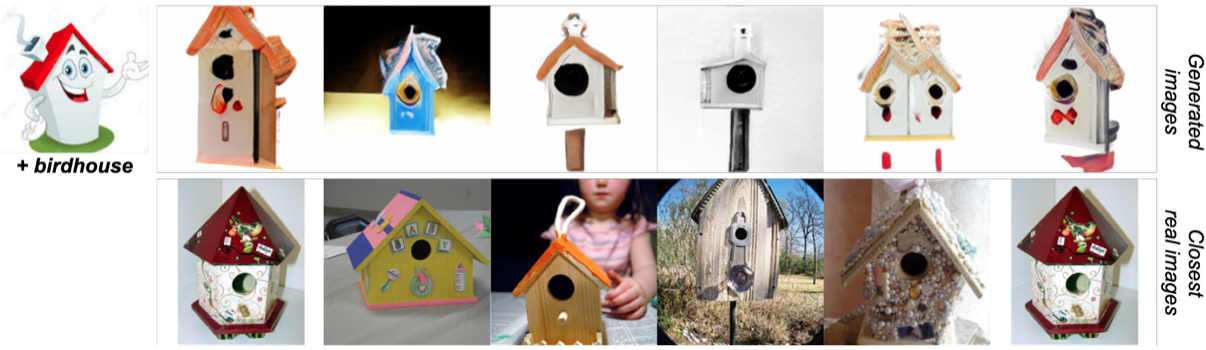}
\end{subfigure}
\caption{Qualitative off-the-shelf transfer results in \myres{256}, using a class-conditional \ours trained on \ImNet and conditioning on unseen instances from other datasets and a class label. The instances come from the following datasets (top to bottom): COCO-Stuff, Cityscapes, PACS (cartoons).
On the left, a data sample $\mathbf{x}_i$ is depicted, used to obtain the instance features $\mathbf{h}_i\,=\,f_\theta(\mathbf{x}_i)$. Next to the data samples, generated images conditioning on $\mathbf{h}_i$ and a class label (under the data samples) are displayed. Just below the generated images, the closest image in the \ImNet training set are shown (Euclidean distance in the feature space of $f_\theta$).
}
\label{fig:cc_icgan_transfer_app}
\end{figure}

\paragraph{Off-the-shelf transfer results for class-conditional \ours.} In Figure~\ref{fig:cc_icgan_transfer_app}, we show additional results when transferring a class-conditional \ours with a BigGAN backbone trained on \ImNet to other datasets, using an \ImNet class label but an unseen instance. We are able to generate camels in the grass by conditioning on an image of a cow in the grass from COCO-Stuff, we show generated images with a zebra in an urban environment by conditioning on a Cityscapes instance, and we generate cartoon-ish birdhouses by conditioning on a PACS cartoon house instance. This highlights the ability of class-conditional \ours to transfer to other datasets \emph{styles} while maintaining the class label semantics.

\section{Class balancing in \ImNet-LT}
\label{app:class_balacing}
We experimented with class balancing for BigGAN in the \ImNet-LT dataset.
In Table~\ref{table:imagenet_lt_balancing}, we compare (1) \textbf{BigGAN}, where both the class distribution for the generator and the data for the discriminator are long-tailed; (2) \textbf{BigGAN (CB)}, a class-balanced version of BigGAN, where the generator samples class labels from a uniform distribution and the samples fed to the discriminator are also class-balanced; and (3) \textbf{BigGAN (T\,=\,2)} where the class distribution is balanced with a softmax temperature of T\,=\,2 providing a middle ground between the long-tail and the uniform distributions. In the latter case, the probability to sample class $c$ (with a frequency $f_c$ in the long-tailed training set) during training is given by $p_c\,=\,\text{softmax}(T^{-1} \ln f_c)$.

Interestingly, balancing the class distribution (either with uniform class distribution or with T=2) harmed performance in all cases except for the validation Inception Score. We hypothesize that oversampling rare classes, and thus the few images therein, may result in overfitting for the discriminator, leading to low quality image generations. 

\begin{table}[h]
\small
\centering
\caption{ImageNet-LT quantitative results for different class balancing techniques.
"t.": training and "v.": validation.}
 \begin{tabular}{@{}lc|cc|ccc@{}}
\toprule
 & \textbf{Res.} & $\downarrow$\textbf{t. FID} & $\uparrow$\textbf{t. IS} & $\downarrow$\textbf{v. FID} & $\downarrow$\textbf{v. [many/med/few] FID} & $\uparrow$\textbf{v. IS} \\  \midrule
BigGAN & 64 & \textbf{27.6} $\pm$ 0.1 & \textbf{18.1} $\pm$ 0.2 & \textbf{28.1} $\pm$ 0.1  & \textbf{28.8/32.8/48.4} $\pm$ 0.2 & 16.0 $\pm$ 0.1  \\
BigGAN (CB) & 64 & 62.1 $\pm$ 0.1 & 10.7 $\pm$ 0.2 & 56.2 $\pm$ 0.1 & 62.2/59.7/74.7 $\pm$ 0.2 & 11.0 $\pm$ 0.0  \\
BigGAN (T\,=\,2) & 64 & 30.6 $\pm$ 0.1 & 16.8 $\pm$ 0.1 & 29.2 $\pm$ 0.1  & 30.9/33.3/49.5 $\pm$ 0.2 & \textbf{16.4} $\pm$ 0.1 \\
\bottomrule
\end{tabular}

 \label{table:imagenet_lt_balancing}
 \end{table}

\section{Choice of feature extractor}
\label{app:feature_extractor}

We study the choice of the feature extractor used to obtain instance features in Table~\ref{table:ablation_knn_feat}. We compare results using an \ours with a BigGAN backbone when coupling it with a ResNet50 feature extractor trained with either self supervision (SwAV) or with supervision for classification purposes (RN50) on \ImNet dataset. Results highlight similar \ours performances for both feature extractors, suggesting that the choice of feature extractor that does not greatly impact the performance of our method when leveraging unlabeled datasets. Given that for the unlabeled scenario we assume no labels are available, we use the SwAV feature extractor. However, in the class-conditional case, we observe that the \ours coupled with a RN50 feature extractor surpasses \ours coupled with a SwAV feature extractor. Therefore, we choose the RN50 feature extractor for the class-conditional experiments. For \ImNet-LT, we transfer these findings and use a RN50 trained on \ImNet-LT as feature extractor, assuming we do not have access to the entire \ImNet dataset and its labels.
 
\begin{table}[h]
\footnotesize 
\centering
\caption{Feature extractor impact with SwAV (ResNet50 trained with a self-supervised approach) and RN50 (ResNet50 trained for the classification task in ImageNet). Experiments performed in \myres{64} ImageNet, using $1,\!000$ training instance features at test time, selected with k-means.}
 \begin{tabular}{@{}lc@{}}
\toprule
  & $\downarrow$\textbf{FID} \\ 
 \midrule
\ours + SwAV & 11.7 $\pm$ 0.1 \\ 
\ours + RN50  & \textbf{11.3} $\pm$ 0.1 \\ 
\midrule
Class-conditional \ours + SwAV & 9.9 $\pm$ 0.1  \\ 
Class-conditional \ours + RN50  & \textbf{8.5} $\pm$ 0.0 \\
\bottomrule
\end{tabular}
 \label{table:ablation_knn_feat}
 \end{table}

\section{Number of conditioning instance features at train time}
\label{app:num_cond_train}
To demonstrate that using many fine-grained overlapping partitions results in better performance than using a few coarser partitions, we trained \ours with a BigGAN backbone by conditioning on all 1.2M training instance features at training time in \ImNet and a neighborhood size of $k\,=\,50$, and compared it quantitatively with an \ours trained by conditioning on only $1,\!000$ instance features at training time. In this case, we extend the neighborhood size to $k\,=\,1,\!200$ to better cover the training distribution~\footnote{Note that this setup resembles the class partition in \ImNet, where $1,\!000$ classes contain approximately $1,\!200$ images each.}. Note that using $k\,=\,50$ would result in using at most $50,\!000$ training samples during training, an unfair comparison. The $1,\!000$ instance features are selected by applying k-means to the entire \ImNet training set. We then use the same instances to generate images. Results are presented in Table~\ref{table:ablation_num_instances_train} and emphasize the importance of training with all available instances, which results in significantly better FID and IS presumably due to the increased number of conditionings and their smaller neighborhoods.

\begin{table}[h]
\footnotesize 
\centering
\caption{Comparison between training \ours (BigGAN backbone) using only $1,\!000$ conditioning instance features (selected with k-means) or all training instance features during training, in \ours \myres{64}. At test time, we condition \ours on $1,\!000$ training instance features, selected with k-means.}
 \begin{tabular}{@{}lcc@{}}
\toprule
 \textbf{Method}
&  $\downarrow$\textbf{FID} & $\uparrow$\textbf{IS} \\  \midrule
\ours ($k\,=\,50$, trained with all 1.2M conditionings)  & \textbf{11.7} $\pm$ 0.1 & \textbf{21.6} $\pm$ 0.1 \\ 
\ours ($k\,=\,1,\!200$, trained with only $1,\!000$ conditionings)  & 24.8 $\pm$ 0.1 & 16.4 $\pm$ 0.1 \\ 
\bottomrule
\end{tabular}
 \label{table:ablation_num_instances_train}
 \end{table}

\section{Matching storage requirements for \ours and unconditional models}
\label{app:fair_comparison}
We hypothesize that the good performance of \ours on \ImNet and COCO-Stuff can not solely be attributed to the slight increase of parameters and the extra memory required to store the instance features used at test time, but also to the \ours design, including the finegrained overlapping partitions and the instance conditionings. To test for this hypothesis, we performed experiments with the unconditional BigGAN baseline on \ImNet and COCO-Stuff, training it by setting all labels in the training set to zero, following~\cite{pmlr-v97-lucic19a,noroozi2020self}, and increasing the generator capacity such that it matches the \ours storage requirements. In particular, we not only endowed the unconditional BigGAN with additional parameters to compensate for the capacity mismatch, but also for the instances required by \ours. Moreover, we performed analogous experiments for the unconditional StyleGAN2 in COCO-Stuff.

\paragraph{\ImNet.} Given its instance conditioning, the \ours (BigGAN backbone) generator introduces an additional 4.5M parameters when compared to the unconditional BigGAN generator. Moreover, \ours requires an extra 8MB to store the $1,\!000$ instance features used at inference time. This 8MB can be translated into roughly 2M parameters\footnote{We store both parameters and instance features as float32.}. 
Therefore, we compensate for this additional storage in \ours by increasing the unconditional BigGAN capacity by expanding the width of both generator and discriminator hidden layers. We follow the practice in~\cite{brock2018large}, where the generator and discriminator's width are changed together. 
The resulting unconditional BigGAN baseline generator has an additional 6.5M parameters. 
Results are reported in Table~\ref{table:ablation_fair_capacity_imagenet}, showing that adding extra capacity to the unconditional BigGAN leads to an improvement in terms of FID and IS. However, \ours still exhibits significantly better FID and IS, highlighting that the improvements cannot be solely attributed to the increase in parameters nor instance feature storage requirements. 

\paragraph{COCO-Stuff.} Similarly, \ours (BigGAN backbone) trained on COCO-Stuff requires 4M additional parameters on top of the extra storage required by the $1,\!000$ stored instance features (8MB again translated into roughly 2M parameters). Therefore, we add 6M extra parameters to the unconditional BigGAN generator. In the case of \ours with a StyleGAN2 backbone, the instance feature conditionings constitute 1M additional parameters. We therefore increase the capacity of the unconditional StyleGAN2 model by 3M to match the storage requirements of \ours (StyleGAN2 backbone). The results are presented in Table~\ref{table:ablation_fair_capacity_coco}, where it is shown that both the unconditional BigGAN and unconditional StyleGAN2 do not take advantage of the additional capacity and achieve poorer performance than the model with lower capacity, possibly due to overfitting. When compared to \ours, the results match the findings in the \ImNet dataset: \ours exhibits lower FID when using BigGAN or StyleGAN2 backbones, compared to their respective unconditional models with the same storage requirements, further highlighting that \ours effectively benefits from its design, finegrained overlapping partitions, and instance conditionings.

\begin{table}[h]
\footnotesize 
\centering
\caption{Comparing \ours with the unconditional counterparts of BigGAN on \myres{64} \ImNet with the same storage requirements. 
Storage-G counts the storage required for the generator, Storage-I the storage required for the training instance features, and Storage-All is the sum of both generator and instance features required storage. FID and IS scores are computed using Pytorch code.}
 \begin{tabular}{@{}lrrrrcc@{}}
\toprule
 \textbf{Method}
&  \textbf{\#prms.} & \textbf{Storage-G} &\textbf{Storage-I} & \textbf{Storage-All}  & $\downarrow$\textbf{FID} & $\uparrow$\textbf{IS} \\ 
\midrule
Unconditional BigGAN & 32.5M & 124MB & 0MB & 124MB & 30.0 $\pm$ 0.1 & 12.1 $\pm$ 0.1 \\
Unconditional BigGAN & 39M & 149MB & 0MB & 149MB & 16.9 $\pm$ 0.0 & 14.6 $\pm$ 0.1 \\
\ours (BigGAN) & 37M & 141MB & 8MB & 149MB & \textbf{10.4} $\pm$ 0.1 & \textbf{21.9} $\pm$ 0.1 \\ 
\bottomrule
\end{tabular}

 \label{table:ablation_fair_capacity_imagenet}
 \end{table}
 
\begin{table}[h]
\footnotesize 
\centering
\caption{Comparing \ours with the unconditional counterparts on \myres{128} COCO-Stuff, with the same storage requirements. 
Storage-G counts the storage required for the generator, Storage-I the storage required for the training instance features, and Storage-All is the sum of both generator and instance features required storage.
}
 \begin{tabular}{@{}lrrrrcc@{}}
\toprule
 & \textbf{\#prms.} & \textbf{Storage-G} &\textbf{Storage-I} & \textbf{Storage-All} &  \multicolumn{2}{c}{$\downarrow$\textbf{FID}} \\ 
  & & & & &  \textbf{train} &  \textbf{eval} \\ 
   \midrule
Unconditional BigGAN & 18M & 68MB & 0MB & 68MB & 17.9 $\pm$ 0.1 & 46.9 $\pm$ 0.5 \\ 
Unconditional BigGAN & 24M & 92MB & 0MB & 92MB & 28.8 $\pm$ 0.1 & 58.1 $\pm$ 0.5 \\
\ours (BigGAN) & 22M & 84MB & 8MB & 92MB & \textbf{16.8} $\pm$ 0.1 & \textbf{44.9} $\pm$ 0.5 \\ 
\midrule
Unconditional StyleGAN2 & 23M & 88MB & 0MB & 88MB & 8.8 $\pm$ 0.1 & 37.8 $\pm$ 0.2 \\
Unconditional StyleGAN2 & 26M & 100MB & 0MB & 100MB & 9.4 $\pm$ 0.0 & 38.4 $\pm$ 0.3\\
\ours (StyleGAN2) & 24M & 92MB & 8MB & 100MB & \textbf{8.7} $\pm$ 0.0 & \textbf{35.8} $\pm$ 0.1\\
\bottomrule
\end{tabular}
 \label{table:ablation_fair_capacity_coco}
 \end{table}

\section{Additional neighborhood size impact studies}
\label{app:k_size}
We additionally study the impact of the neighborhood size for \ImNet-LT in Table~\ref{table:ablation_lt} and in COCO-Stuff in Table~\ref{table:ablation_coco}, showing that in both cases, \ours with a BigGAN backbone and $k\,=\,5$ achieves the best FID and IS metrics. The choice of a lower neighborhood size $k\,=\,5$ than in the \ImNet case ($k\,=\,50$) could suggest that the number of semantically similar neighboring samples is smaller for these two datasets. This wouldn't be completely surprising given that these two datasets are considerably smaller than \ImNet. Increasing the value of $k$ in COCO-Stuff and \ImNet-LT would potentially gather samples with different semantics within a neighborhood, which could potentially harm the training and therefore the generated images quality.
 
\begin{figure}[h]
\footnotesize 
\centering
\begin{subfigure}{0.1\textwidth}
 \centering
\includegraphics[width=\textwidth]{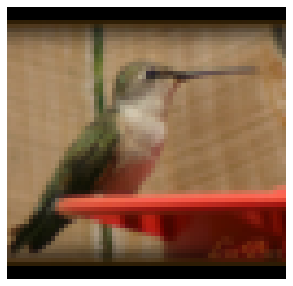}
\caption{$\mathbf{x}_i$}
\label{subfig:icgan_unlabeled_input1}
\end{subfigure}
\begin{subfigure}{0.31\textwidth}
 \centering
\includegraphics[width=\textwidth]{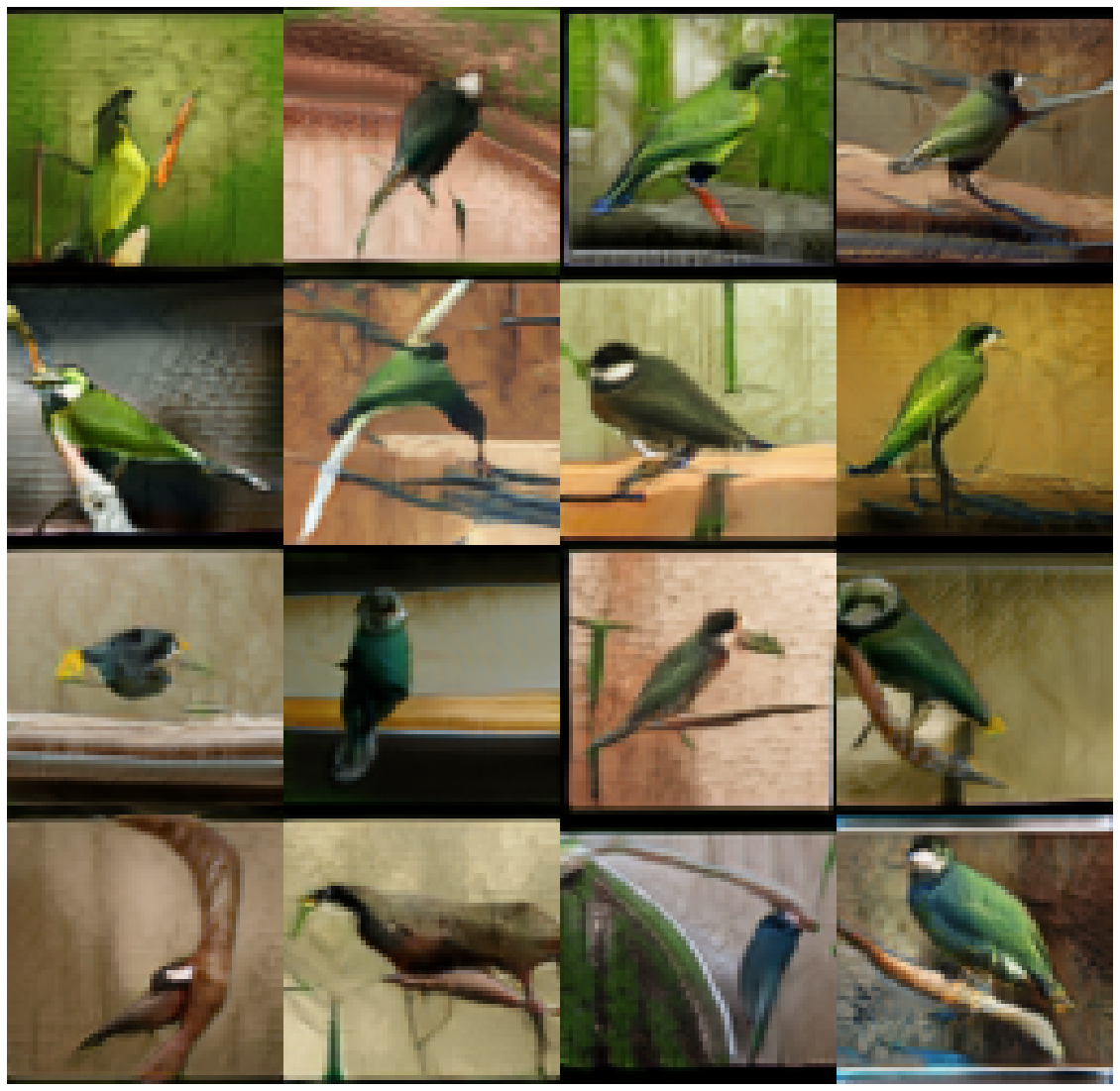}
\caption{\ours trained with $k\!=\!5$}
\label{subfig:bird_k5}
\end{subfigure}
\begin{subfigure}{0.31\textwidth}
 \centering
\includegraphics[width=\textwidth]{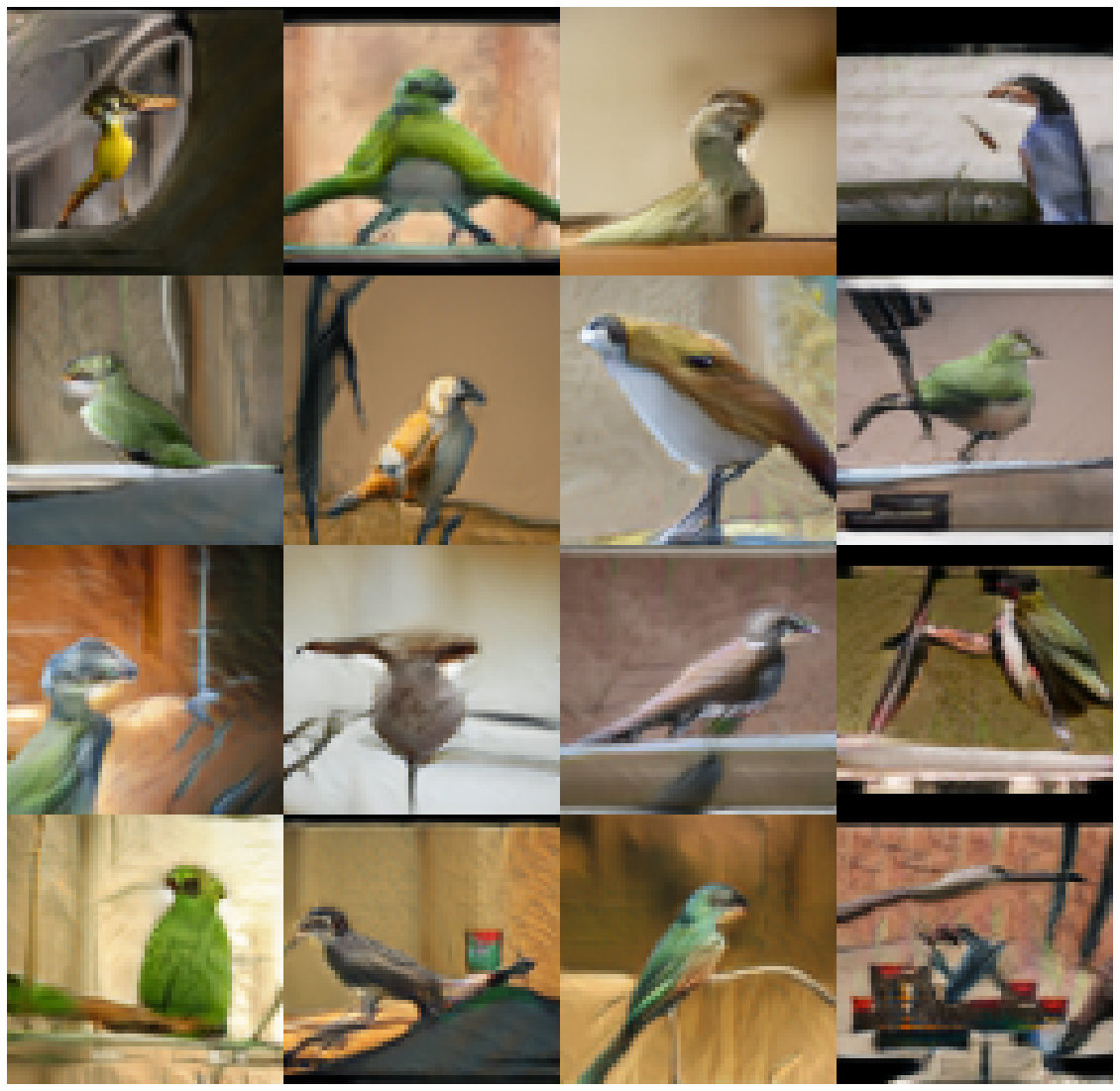}
\caption{\ours trained with $k\!=\!20$}
\label{subfig:bird_k20}
\end{subfigure}
\begin{subfigure}{0.31\textwidth}
 \centering
\includegraphics[width=\textwidth]{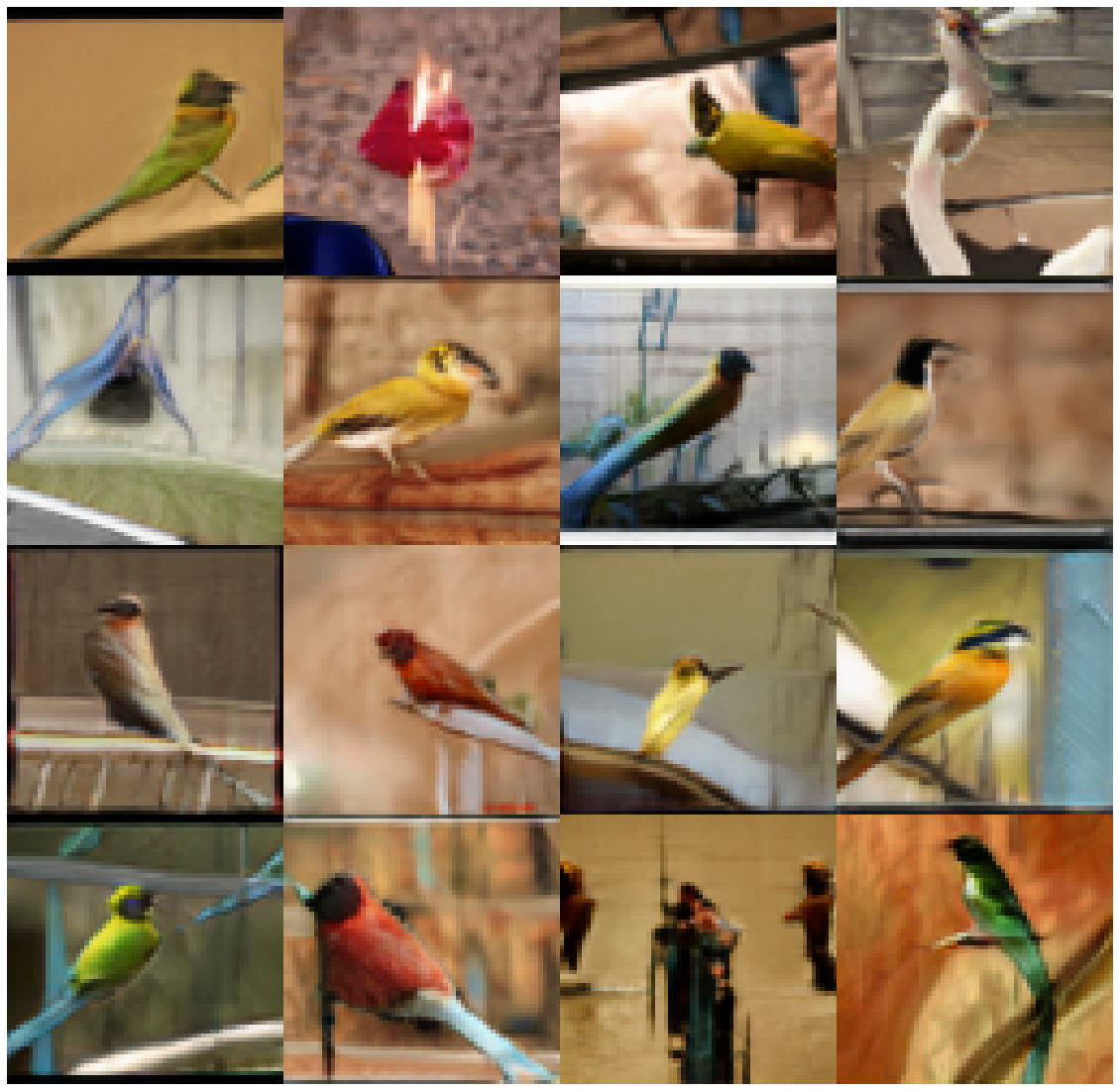}
\caption{\ours trained with $k\!=\!50$}
\label{subfig:bird_k50}
\end{subfigure}
\begin{subfigure}{0.31\textwidth}
 \centering
\includegraphics[width=\textwidth]{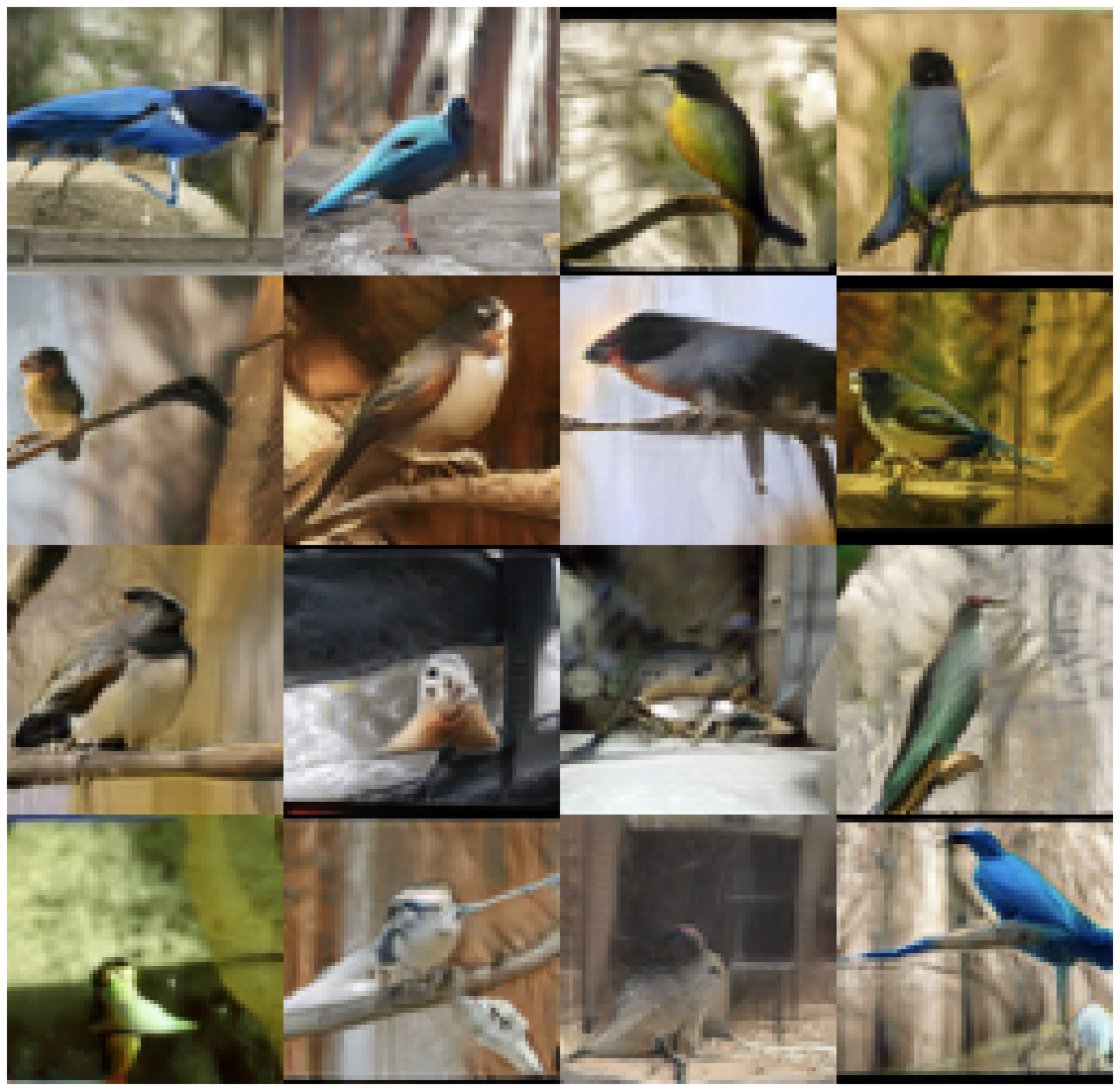}
\caption{\ours trained with $k\!=\!100$}
\label{subfig:bird_k100}
\end{subfigure}
\begin{subfigure}{0.31\textwidth}
 \centering
\includegraphics[width=\textwidth]{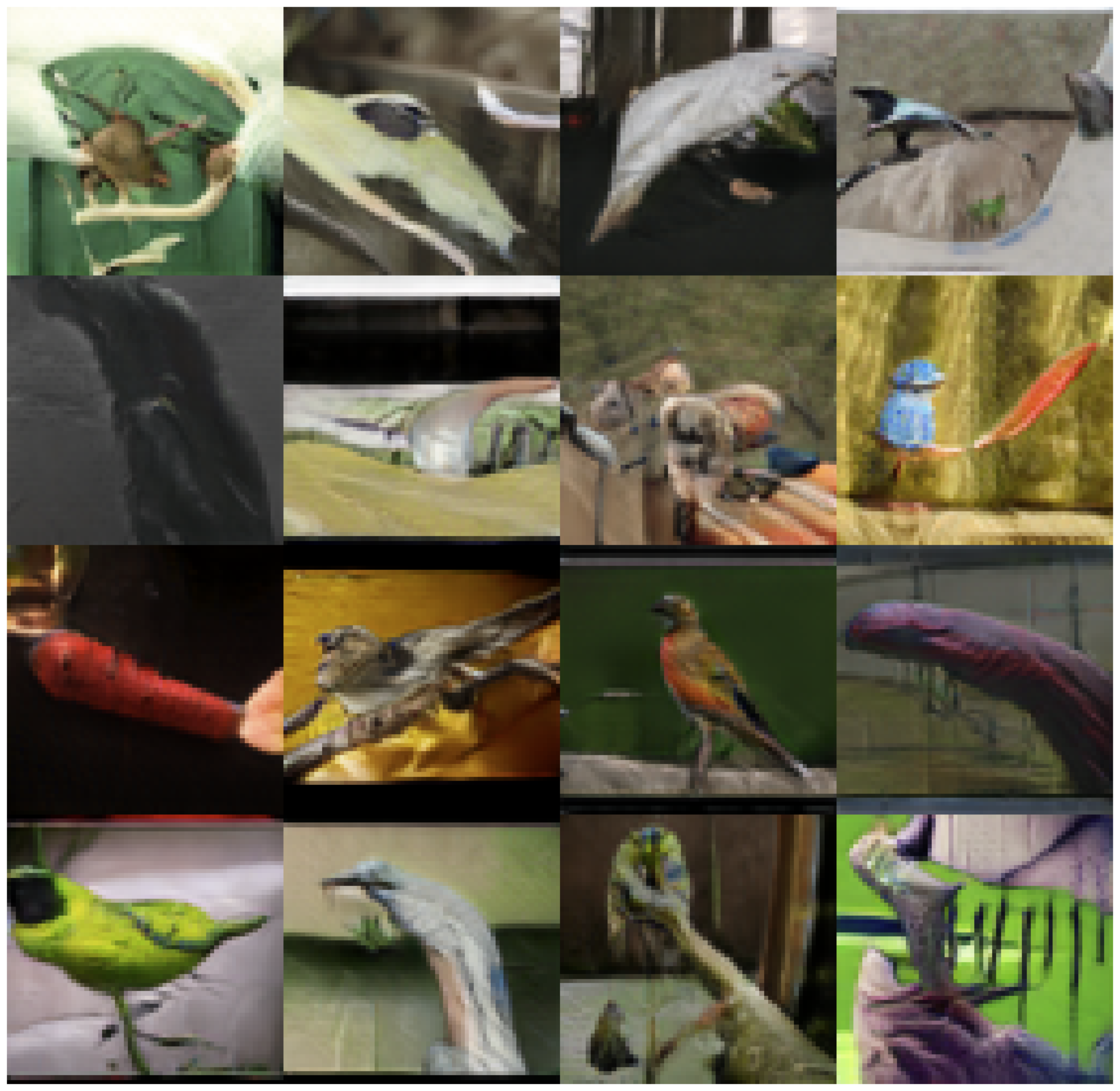}
\caption{\ours trained with $k\!
=\!500$}
\label{subfig:bird_k500}
\end{subfigure}
\caption{Qualitative results in \myres{64} unlabeled \ImNet when training \ours (BigGAN backbone) with different neighborhood sizes $k$. (a) Data samples $\mathbf{x}_i$ used to obtain the instance features $\mathbf{h}_i\,=\,f_\theta(\mathbf{x}_i)$.  (b-f) Generated images with \ours (BigGAN backbone), sampling different noise vectors, for different neighborhood sizes $k$ used during training.}
\label{fig:icgan_k_effect}
\end{figure}

Finally, in Figure~\ref{fig:icgan_k_effect}, we qualitatively show generated images in \ImNet when using an \ours trained with varying neighborhood sizes. The findings further support the ones presented in Subsection~\ref{sec:ablation}, showing that smaller neighborhoods result in generated images with less diversity, while bigger neighborhood sizes, for example $k\,=\,500$ result in more varied but lower quality generations, supporting that $k$ controls the smoothing effect.

\begin{table}[h]
\footnotesize 
\caption{Impact of the number of neighbors ($k$) used to train class-conditional \ours (BigGAN backbone) in ImageNet-LT \myres{64}. Reported results in ImageNet validation set. As a feature extractor, a ResNet50 is trained as a classifier on the same dataset is used. 50k instance features are sampled from the training set.}
\centering
\begin{tabular}{@{}lcc@{}}
\toprule
 & $\downarrow$\textbf{FID} & $\uparrow$\textbf{IS}  \\ \midrule
$k$\,=\,5 & \textbf{23.4} $\pm$ 0.1 & \textbf{17.6} $\pm$ 0.1 \\ 
$k$\,=\,20 & 24.1 $\pm$ 0.1 & 16.8 $\pm$ 0.1 \\ 
$k$\,=\,50 & 24.1 $\pm$ 0.1 & 16.7 $\pm$ 0.1 \\ 
$k$\,=\,100 & 25.6 $\pm$ 0.1 & 16.3 $\pm$ 0.1 \\
$k$\,=\,500 & 27.1 $\pm$ 0.1 & 15.3 $\pm$ 0.1 \\ 
\bottomrule
\end{tabular}
\label{table:ablation_lt}
\end{table}

\begin{table}[h]
\footnotesize 
\caption{Impact of the number of neighbors ($k$) used to train \ours (BigGAN backbone) on COCO-Stuff \myres{128}. Reported results on COCO-Stuff evaluation set. As a feature extractor, a ResNet50 trained with self-supervision (SwAV) is used.}
\centering
 \begin{tabular}{@{}lcc@{}}
\toprule
 & $\downarrow$\textbf{FID} \\ \midrule
$k$\,=\,5 & \textbf{44.9} $\pm$ 0.5 \\ 
$k$\,=\,20 & 46.8 $\pm$ 0.3 \\ 
$k$\,=\,50 & 45.8 $\pm$ 0.4 \\ 
$k$\,=\,100 & 48.4 $\pm$ 0.3 \\ 
$k$\,=\,500 & 48.3 $\pm$ 0.5 \\
\bottomrule
\end{tabular}
  \label{table:ablation_coco}
 \end{table}

\end{document}